\documentclass[5p, number, sort&compress]{elsarticle}

\usepackage{amsmath}
\usepackage{enumitem}

\usepackage{booktabs}
\usepackage[flushleft]{threeparttable}
\usepackage{adjustbox}

\usepackage{graphicx}
\usepackage{environ}
\usepackage[utf8]{inputenc} 
\usepackage[T1]{fontenc}    
\usepackage[hidelinks]{hyperref}
\usepackage{url}            
\usepackage{amsfonts}       
\usepackage{xcolor}         
\usepackage{amsthm}
\usepackage{amssymb}
\usepackage{algorithm}
\usepackage[noend]{algpseudocode}
\algrenewcommand\algorithmicdo{}
\usepackage{tikz}
\usepackage{pgfplots}
\pgfplotsset{compat=1.18}
\usepackage{tabularx}
\usepackage{tabulary}
\usepackage{multirow}

\usepackage{setspace}

\usepackage{caption}
\usepackage{cleveref}
\usepackage{anyfontsize}

\usepackage{subcaption}
\usepackage[separate-uncertainty=true]{siunitx}

\usetikzlibrary{backgrounds}
\tikzset{font={\fontsize{9pt}{10}\selectfont}}
\usepackage{cases}
\usepackage[skins]{tcolorbox}
\usetikzlibrary{external,matrix,positioning}
\usepgfplotslibrary{groupplots,statistics,patchplots,fillbetween,ternary}

\algnewcommand\algorithmicforeach{\textbf{for each}}
\algdef{S}[FOR]{ForEach}[1]{\algorithmicforeach\ #1\ \algorithmicdo}

\usepackage{xifthen} % for \ifthenelse{}{}{}
\usepackage{etoolbox} % for other programming stuff
\newcommand*{\ifempty}[3]{\ifthenelse{\isempty{#1}}{#2}{#3}}

\makeatletter
\newcommand{\algcolor}[2]{%
  \hskip-\ALG@thistlm\colorbox{#1}{\parbox{\dimexpr\linewidth-2\fboxsep}{\hskip\ALG@thistlm\relax #2}}%
}

\makeatother

\newcommand{\pctapf}{PC-TAPF} % precedence-constrained multi-agent path finding
\newcommand{\pcta}{PC-TA} % precedence-constrained multi-agent path finding
\newcommand{\numrobots}{n}

\newcommand{\numtasks}{m}
\newcommand{\StartTime}[1][]{t_{#1}^{0}}
\newcommand{\CompletionTime}[1][]{t_{#1}^{F}}

\newcommand{\ProjectSchedule}{S}

\newcommand{\ScheduleVertices}{V_{\ProjectSchedule}}
\newcommand{\ScheduleEdges}{E_{\ProjectSchedule}}
\newcommand{\vtx}{v}
\newcommand{\vtxONE}{v}
\newcommand{\vtxTWO}{u}

\newcommand{\RequiredPredecessors}{\small \textsc{RequiredPred}}
\newcommand{\RequiredSuccessors}{\small \textsc{RequiredSucc}}
\newcommand{\EligiblePredecessors}{\small \textsc{EligiblePred}}
\newcommand{\EligibleSuccessors}{\small \textsc{EligibleSucc}}
\newcommand{\EligibleEdges}{\small \textsc{EligibleEdges}}

\newcommand{\processtime}[1][]{\Delta t_{#1}}

\newcommand{\adjacencyMatrix}[2]{X_{#1\ifempty{#1}{}{,}#2}}
\newcommand{\adjMtx}{\adjacencyMatrix}

\newcommand{\milp}{\textsc{milp}}

\newcommand{\idxA}{\vtxONE}
\newcommand{\idxB}{\vtxTWO}

\providetoggle{tightmilp}
\settoggle{tightmilp}{false}

\newcommand{\TerminalProjectNodes}{\small \textsc{TerminalProjectNodes}}
\newcommand{\eightspans}{\span\span\span\span\span\span\span\span}

\newcommand{\reals}{{\mathbb R}}

\newcommand{\booleans}{{\mbox{\bf B}}}
\newcommand{\vect}[1]{\mathbf{#1}}
\newcommand{\mat}[1]{\vect{#1}}

\newcommand{\ra}[1]{\renewcommand{\arraystretch}{#1}}

\newcommand{\LDraw}{LDraw$^{\text{\texttrademark}}$}
\newcommand{\LEGO}{LEGO$^{\text{\textregistered}}$}

\newcommand{\greedy}{\textsc{Greedy}}

\newcommand{\worldframe}{W}

\newcommand{\vmaxrobot}{v_{\textsc{max}}}
\newcommand{\vminrobot}{v_{\textsc{min}}}
\newcommand{\posvec}{\mat{x}}
\newcommand{\posvel}{\dot{\posvec}}

\newcommand{\rectvolume}{\textsc{r\_volume}}
\newcommand{\velvolumefactor}{\textsc{v\_factor}}
\newcommand{\robotradius}{r}
\newcommand{\minrobots}{\underline{n}}

\newcommand{\bestidxs}{best\_idxs}
\newcommand{\candidateidxs}{idxs}
\newcommand{\idxneighbors}[1]{\textsc{neighbors}(#1)}
\newcommand{\idxscore}[1]{\textsc{score}(#1)}
\newcommand{\updatedflag}{updated}
\newcommand{\CompRadius}[1]{r_{#1}}

\newcommand{\AssemblyRadius}{R}
\newcommand{\CompAngle}[1]{\theta_{#1}}
\newcommand{\DesiredCompAngle}[1]{\hat{\theta}_{#1}}
\newcommand{\HalfWidth}[1]{\Delta_{#1}}
\newcommand{\ObjectStart}{\textsc{ObjectStart}}
\newcommand{\RobotStart}{\textsc{RobotStart}}
\newcommand{\RobotGo}{\textsc{RobotGo}}
\newcommand{\AssemblyStart}{\textsc{AssemblyStart}}
\newcommand{\OpenBuildStep}{\textsc{OpenBuildStep}}
\newcommand{\CloseBuildStep}{\textsc{CloseBuildStep}}
\newcommand{\FormTransportUnit}{\textsc{FormTransportUnit}}
\newcommand{\TransportUnitGo}{\textsc{TransportUnitGo}}
\newcommand{\DepositCargo}{\textsc{DepositCargo}}
\newcommand{\LiftIntoPlace}{\textsc{LiftIntoPlace}}
\newcommand{\AssemblyComplete}{\textsc{AssemblyComplete}}
\newcommand{\ProjectComplete}{\textsc{ProjectComplete}}

\newcommand{\AssignmentMILP}{\textsc{AssignmentMILP}}
\newcommand{\AdjacencyMILP}{\textsc{AdjacencyMILP}}
\newcommand{\SparseAdjacencyMILP}{\textsc{SparseAdjacencyMILP}}
\newcommand{\greedyLEGO}{\greedy{}-\textsc{PCCF}}
\newcommand{\x}{\posvec}
\newcommand{\vel}{\mat{v}}

\newcommand{\vnom}{\vel_{\textnormal{nominal}}}
\newcommand{\vpref}{\vel_{\textnormal{preferred}}}
\newcommand{\vcom}{\vel_{\textnormal{commanded}}}
\newcommand{\TangentBugPolicy}{\textsc{TangentBugPolicy}}
\newcommand{\DispersionProtocol}{\textsc{DispersionProtocol}}
\newcommand{\RVO}{\textsc{RVO}}
\newcommand{\waypoint}{waypoint}
\newcommand{\target}{target}
\newcommand{\buggoal}{goal}
\newcommand{\bugpos}{pos}
\newcommand{\bugline}{\bugpos \rightarrow \buggoal}
\newcommand{\bugradius}{\textsc{planning\_radius}}
\newcommand{\agentRadius}{r}
\newcommand{\bufferRadius}{R}
\newcommand{\maxbufferRadius}{\bufferRadius_{\textsc{max}}}
\newcommand{\bufferMultiplier}{c}
\newcommand{\potentialfield}{F}

\newcommand{\rvoagent}{agent}

\newcommand{\incfig}[1]{\raisebox{-0.25\totalheight}{\includegraphics[scale=0.85]{gs_#1}}}
\newcommand{\ObjectStartN}{\incfig{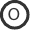}}
\newcommand{\RobotStartN}{\incfig{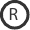}}
\newcommand{\RobotGoN}{\incfig{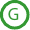}}
\newcommand{\AssemblyStartN}{\incfig{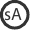}}
\newcommand{\OpenBuildStepN}{\incfig{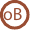}}
\newcommand{\FormTransportUnitN}{\incfig{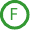}}
\newcommand{\TransportUnitGoN}{\incfig{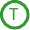}}
\newcommand{\DepositCargoN}{\incfig{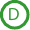}}
\newcommand{\LiftIntoPlaceN}{\incfig{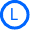}}
\newcommand{\CloseBuildStepN}{\incfig{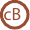}}
\newcommand{\AssemblyCompleteN}{\incfig{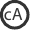}}
\newcommand{\ProjectCompleteN}{\incfig{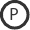}}

\tikzset{level plate/.style={%
      thick,
      fill=black!20,
      text depth=0pt,
      rounded corners,
      inner ysep=4pt,
      inner xsep=4pt,
  },
  empty level plate/.style={%
      text depth=0pt,
      anchor=west,
      align=center,
      minimum width=0cm,
      minimum height=0cm,
      inner ysep=0pt,
      inner xsep=0pt
  },
  data block/.style={%
      rounded corners,
      text depth=0pt,
      thick,
      minimum width=0cm,
      minimum height=0cm,
      fill=blue!35,
      thick,
      inner ysep=5pt,
      inner xsep=5pt
  },
  empty data block/.style={%
      minimum height=0cm,
      minimum width=0cm,
      text depth=0pt,
      inner ysep=0pt,
      inner xsep=0pt
  },
  function block/.style={%
      thick,
      minimum width=0cm,
      minimum height=0cm,
      text depth=0pt,
      fill=red!40,
      rounded corners,
      inner ysep=5pt,
      inner xsep=5pt,
      minimum width=0.5cm
  },
  empty function block/.style={%
      minimum width=0cm,
      minimum height=0cm,
      text depth=0pt,
      rounded corners,
      inner sep=0pt
  },
  intermediate output block/.style={%
      minimum width=0cm,
      minimum height=0cm,
      text depth=0pt,
      rounded corners,
      inner sep=0pt
  },
  thick edge/.style={%
    >=latex,
    ->,
    thick,
    shorten >=1pt,
    shorten <=1pt,
  }
}
\newcommand{\objectcolor}{orange!20}
\newcommand{\robotcolor}{green!20}
\newcommand{\cleanupbotcolor}{purple!20}
\newcommand{\actioncolor}{blue!20}
\newcommand{\opcolor}{red!20}
\tikzset{BaseScheduleNode/.style={inner sep=3pt, outer sep=1pt, align=center,minimum size=0pt}}
\tikzset{TaskNode/.style={BaseScheduleNode,shape=circle, text=black!, fill=black!20, opacity=1}}
\tikzset{RobotNode/.style={BaseScheduleNode,shape=circle, text=black!, fill=\robotcolor, opacity=1}}
\tikzset{CleanUpBotNode/.style={BaseScheduleNode,shape=circle, text=black!, fill=\cleanupbotcolor, opacity=1}}
\tikzset{RobotTerminalNode/.style={BaseScheduleNode,shape=diamond, text=black!, fill=\robotcolor, opacity=1}}
\tikzset{ObjectNode/.style={BaseScheduleNode,shape=circle, text=black!, fill=\objectcolor, opacity=1}}
\tikzset{ObjectTerminalNode/.style={BaseScheduleNode,shape=diamond, text=black!, fill=\objectcolor, opacity=1}}
\tikzset{ActionNode/.style={BaseScheduleNode,shape=rectangle, rounded corners=4pt,text=black!, fill=\actioncolor, opacity=1}}
\tikzset{OpNode/.style={BaseScheduleNode,shape=regular polygon,regular polygon sides=4, rounded corners=0pt, text=black!, fill=\opcolor, opacity=1}}
\tikzset{ScheduleEdge/.style={<-,>=latex, shorten < = 1pt, shorten > = 1pt,draw,color=black,thick}}
\tikzset{BlockArrow/.style={single arrow, fill=red!20, minimum height=1.5em,minimum width=1.5em,
    single arrow head extend=0.15cm, outer sep=0pt}}

\newcounter{overviewcounter}
\newcommand{\compoverviewcounter}[1]{\ifthenelse{\value{overviewcounter}>#1}{black}{white}}

\newcounter{milpEdgesDrawingMode}
\newcounter{showinputsandoutputs}
\newcommand{\printEdgeText}[1]{\ifthenelse{\value{showinputsandoutputs}=1}{#1}{}}

\newcommand{\milpobjectivefontsize}{\normalsize}
\newcommand{\milpfontsize}{\normalsize}

\newcommand{\frozenedgecolor}{black}
\newcommand{\frozentextcolor}{black}
\newcommand{\frozendrawcolor}{none}
\newcommand{\frozenobjectcolor}{\objectcolor}
\newcommand{\frozenrobotcolor}{\robotcolor}
\newcommand{\frozencleanupbotcolor}{\cleanupbotcolor}
\newcommand{\frozenactioncolor}{\actioncolor}
\newcommand{\frozenopcolor}{\opcolor}
\tikzset{FrozenRobotNode/.style={RobotNode,fill=\frozenrobotcolor,draw=\frozendrawcolor,text=\frozentextcolor}}
\tikzset{FrozenCleanUpBotNode/.style={CleanUpBotNode,fill=\frozencleanupbotcolor,draw=\frozendrawcolor,text=\frozentextcolor}}
\tikzset{FrozenObjectNode/.style={ObjectNode,fill=\frozenobjectcolor,draw=\frozendrawcolor,text=\frozentextcolor}}
\tikzset{FrozenActionNode/.style={ActionNode,fill=\frozenactioncolor,draw=\frozendrawcolor,text=\frozentextcolor}}
\tikzset{FrozenOpNode/.style={OpNode,fill=\frozenopcolor,draw=\frozendrawcolor,text=\frozentextcolor}}
\tikzset{FrozenScheduleEdge/.style={ScheduleEdge,color=\frozenedgecolor}}

\journal{Robotics and Autonomous Systems}

\begin{document}

\begin{frontmatter}

\title{Large-Scale Multi-Robot Assembly Planning for Autonomous Manufacturing}

\author[inst1]{Kyle Brown}
\ead{kylejbrown@alumni.stanford.edu}
\author[inst1]{Dylan M. Asmar\corref{cor1}}
\ead{asmar@stanford.edu}
\author[inst2]{Mac Schwager}
\ead{schwager@stanford.edu}
\author[inst1]{Mykel J. Kochenderfer} 
\ead{mykel@stanford.edu}

\cortext[cor1]{Corresponding author}

\affiliation[inst1]{
    organization={Stanford Intelligent Systems Laboratory, Stanford University},
    addressline={496 Lomita Mall}, 
    city={Stanford},
    state={CA},
    postcode={94350}, 
    country={USA}}

\affiliation[inst2]{
    organization={Multi-Robot Systems Lab, Stanford University},
    addressline={496 Lomita Mall}, 
    city={Stanford},
    state={CA},
    postcode={94350}, 
    country={USA}}


\begin{abstract}
  Mobile autonomous robots have the potential to revolutionize manufacturing processes. However, effective employment of large robot fleets in manufacturing requires addressing numerous challenges including the collision-free movement of multiple agents in a shared workspace, effective multi-robot collaboration to manipulate and transport large payloads, complex task allocation due to coupled manufacturing processes, and spatial planning for parallel assembly and transportation of nested subassemblies. In this work, we propose a full algorithmic stack for large-scale multi-robot assembly planning that addresses these challenges and can synthesize construction plans for complex assemblies with thousands of parts in a matter of minutes. Our approach takes in a CAD-like product specification and automatically plans a full-stack assembly procedure for a group of robots to manufacture the product. We propose an algorithmic stack that comprises: (i) an iterative radial layout optimization procedure to define a global staging layout for the manufacturing facility, (ii) a `graph-repair' mixed-integer program formulation and a modified greedy task allocation algorithm to optimally allocate robots and robot sub-teams to assembly and transport tasks, (iii) a geometric heuristic and a hill-climbing algorithm to plan collaborative carrying configurations of robot sub-teams, and (iv) a distributed control policy that enables robots to execute the assembly motion plan without colliding with each other. We also present an open-source multi-robot manufacturing simulator implemented in Julia as a resource to the research community, to test our algorithmic stack and to facilitate multi-robot manufacturing research more broadly: \url{https://github.com/sisl/ConstructionBots.jl}. Our empirical results demonstrate the scalability and effectiveness of our approach by generating plans to manufacture a \LEGO{} model of a Saturn V launch vehicle with $1845$ parts, $306$ subassemblies, and $250$ robots in under three minutes on a standard laptop computer.
\end{abstract}

\end{frontmatter}

\setcounter{footnote}{0}

\section{Introduction} \label{sec:intro}
Consider a flexible factory environment in which a team of mobile robots must collaborate to construct a large assembly from a collection of discrete components. An assembly plan is given to the factory, which provides a tree of assembly operations to iteratively combine components into progressively larger subassemblies until the final assembly is complete. To fulfill the assembly plan, the robotic factory must spatially configure a set of construction stations for the subassemblies, culminating in the final assembled product at a final station. The factory then needs to produce a motion plan for the robots to shuttle parts and subassemblies between the stations to realize the abstract assembly plan.  As individual components are combined into larger and larger subassemblies, the plan must allow robots to collaboratively transport the larger payloads, taking into account the load-carrying capacity of individual robots.  Finally, the robots must avoid collisions with each other as they navigate the environment to collect components and deliver them to the appropriate locations. In this paper, we propose an algorithmic stack to solve these robot planning and coordination problems that are central to multi-robot manufacturing. We also present a multi-robot manufacturing simulator, ConstructionBots.jl, implemented in Julia and open-sourced for the research community to facilitate research in multi-robot manufacturing.

\begin{figure*}[t]
    \centering
    \resizebox{\linewidth}{!}{%
        \input{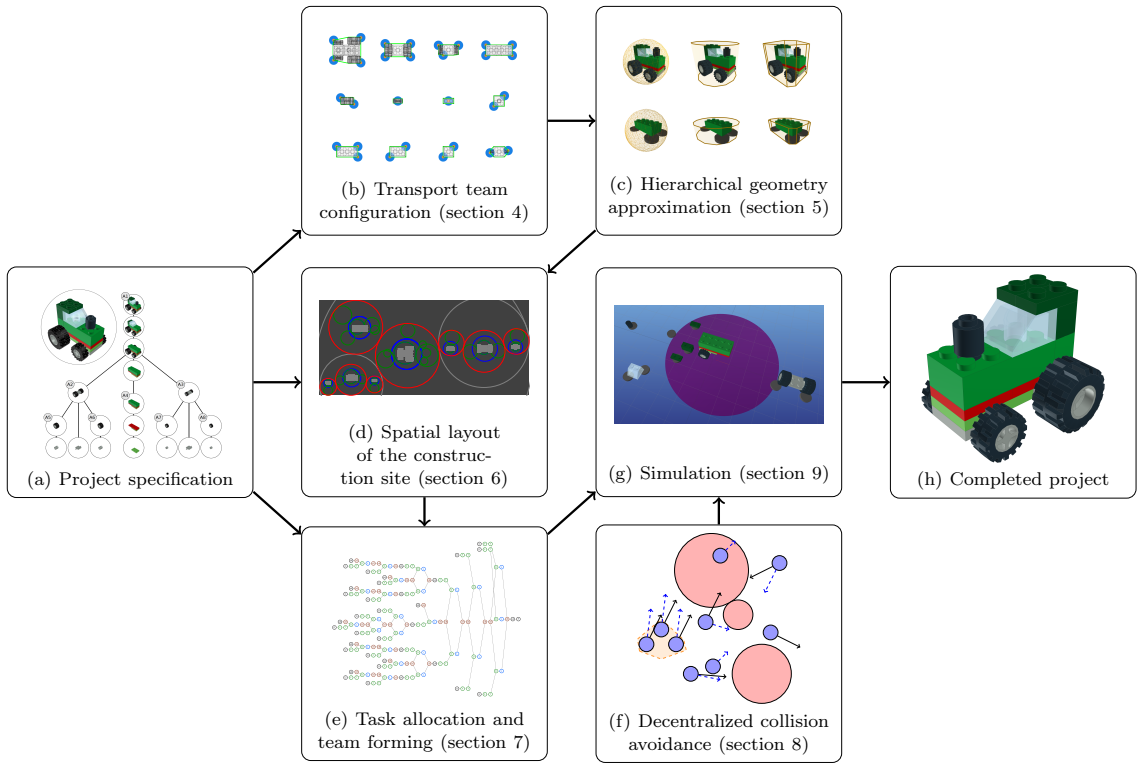}
    }
    \caption{An overview of the proposed multi-robot assembly planning system. (a) Starting with a CAD-like project specification, the process evolves to determine (b) the configuration of transport teams, then calculates a (c) hierarchical geometric approximation of parts and transport units. (d) Based on the geometry, a spatial layout for the construction site is designed. (e) Task allocation and team formations are computed, following which (f) a decentralized strategy ensures collision-free execution. (g) The entire planned procedure is simulated, culminating in the (h) final assembled project. The arrows indicate the sequential flow of the planning process.}
    \label{fig:methods_overview}
\end{figure*}

% Why is it interesting
We consider this multi-robot construction concept as an important facet of \emph{Industry 4.0} \citep{lasi2014industry}, the widely heralded fourth industrial revolution fueled by advances in autonomy, AI, and ubiquitous wireless connectivity. Although modern assembly lines are optimized to produce complex assemblies at high speeds, they are tailored to a specific product. Reconfiguring an assembly line to manufacture a new product or a custom variation on a product, can be a costly, time-consuming, human labor-intensive effort \citep{Mehrabi2000, KOREN2010130}. In contrast, the multi-robot construction system concept described above has the potential to deliver faster, cheaper, more customizable, and more reconfigurable fabrication for a broad range of assemblies.
Such a system would be capable of building any assembly for which (a) the raw materials and subassemblies can be transported by robot teams and (b) the atomic operations required to incorporate each material or subassembly into its parent assembly are supported by the factory tooling. In this work we consider atomic part-to-part fastening operations as existing primitives, hence, we do not present research on the control of contact forces, insertion, screwing, riveting, soldering, welding, etc. Our work is focused on task planning, motion planning, and collaborative teaming. 

% Why hasn't it been done before?
Collaboration between robots introduces additional complexity to the planning and control problem, particularly due to the coordination required to manage shared tasks. However, the computational complexity of multi-robot planning does not necessarily scale directly with the number of robots. Some planning algorithms experience dramatically increasing computational demands as the number of robots grows, typically driven by worst-case exponential complexity or algorithm-specific scaling limitations \citep{Yu_LaValle_2013, Lin2022}. In contrast, other methods scale more favorably by leveraging structure inherent to multi-robot tasks. Understanding these scalability characteristics is important, as existing multi-robot assembly planning approaches often struggle with as system complexity increases.

Various systems have been proposed for ``end-to-end'' multi-robot assembly planning and execution \citep{Knepper2013,Dogar2015}. These approaches perform both high-level task planning, coarse ``transit'' motion planning, and detailed manipulation planning required to fasten assembly parts together (e.g., screwing and riveting). Other approaches focus on the geometric assembly planning task, which amounts to determining in what order, and along what paths, to assemble a given assembly subject to constraints that all components must move into their goal configurations without interfering (i.e., colliding) with other components \citep{Wilson1994,wilson1992geometric,Halperin2000,Culbertson}. \citet{Dogar2019} address the geometric assembly planning process with the added complexity of planning sequences of robot poses to realize the assembly process. Though these works represent significant progress toward the goal of autonomous manufacturing, there are still challenges in scalability, multi-robot coordination, and a system-view integration of the many layers of planning required for this problem. 

In contrast to existing work, our system generally approaches multi-robot assembly planning from a higher level of abstraction. We address task planning and transit planning but abstract away the kino-dynamic details of piecing together assemblies. As such, we are able to focus on larger assemblies than those often handled by these more detailed end-to-end approaches.
We address the following specific problems:
\begin{itemize}
   \item \textbf{Transport team configuration} -- How many robots are needed and how should robots be positioned when transporting a particular payload?
  \item \textbf{Spatial layout of the construction site} -- Where will each assembly be built, and where will the components of those assemblies be delivered?
  \item \textbf{Sequential task allocation and team forming} -- Which robots will collect and deliver which payloads? Since there are generally far more payloads than robots, individual robots generally have to transport multiple payloads in sequence.
  \item \textbf{Collision avoidance with heterogeneous agent geometry and dynamics} -- How must laden and unladen robots and robot teams move, subject to motion constraints that depend on the payload size and team configuration, to avoid collision with other robots and the active construction sites in the environment?
\end{itemize}

We present a proof-of-concept system that can synthesize construction plans for assemblies with thousands of parts in a matter of minutes. This computational efficiency enables rapid iteration and exploration of the design space, allowing users to evaluate trade-offs across different metrics such as makespan, spatial efficiency, and resource requirements before committing to a final assembly layout. Such capability becomes particularly valuable for custom products, one-off prototypes, or layout reconfiguration due to supply chain or operational changes. To illustrate the environment model and the process of synthesizing a construction plan, we introduce a simple ``tractor'' project as a running example. This assembly was defined in LeoCAD and is based on \LEGO{} model 10708, \emph{Green Creativity Box}. The tractor model has a total of $20$ individual pieces, which are organized into one final assembly (the tractor) and seven subassemblies. An overview of our proposed process using the tractor model is provided in \cref{fig:methods_overview}.

We review the related literature in \cref{sec:LEGO_background} and define the environment model in \cref{sec:LEGO_environment}. 
In \cref{sec:configure_transport_units}, we introduce an approach for determining multi-robot carrying configurations for transporting objects and then describe our method for generating the spatial layout of a construction site in \cref{sec:construct_staging_plan}. \Cref{sec:collaborative_task_allocation} introduces our approach to task allocation and team forming and we describe a decentralized strategy for plan execution and collision avoidance in \cref{sec:rvo_route_planning}. \Cref{sec:LEGO_demos} reports on several simulations demonstrating our system on various assemblies and then we provide limitations and future work in our discussion in \cref{sec:LEGO_discussion}.

\section{Related Work}\label{sec:LEGO_background}

Our problem falls under the umbrella of Task and Motion Planning (TAMP) problems, which combine discrete task planning with multi-modal continuous motion planning \citep{Garrett2021}. TAMP is a broad framework that is applicable when one or more robots must both move through a continuous environment and modify the state of objects in the environment.
General TAMP problems may incorporate geometric, kinodynamic, and modal variables and constraints. Our problem setting involves variables and constraints in these categories, and could theoretically be expressed as a generic TAMP problem and solved by a general-purpose TAMP solver. 

For a comprehensive review of TAMP methodologies, readers are encouraged to consult the survey by \citet{guo2023}, which presents a broad spectrum of TAMP strategies. Our methodology aligns with the optimization-based approaches discussed in that survey, using a `graph-repair' mixed-integer linear program formulation to effectively manage task allocation in large-scale manufacturing environments with teams of robots. Our approach begins with a project specification that effectively segments the overall assembly task, thereby alleviating the need for extensive task decomposition. This strategic starting point allows us to develop a method tailored specifically to these demands rather than relying on a general-purpose TAMP solver.

An important element of TAMP is the notion of a kinematic graph, which specifies kinematic constraints between entities in the environment \citep{LavallePlanning2006}. As robots interact with the world, the kinematic graph undergoes \emph{mode switches}, wherein edges are added, removed, or modified. A kinematic graph is our main tool for modeling the transient ``pick up'', ``put down'', and ``lock into parent assembly'' modal switches that occur as robots carry components through the environment and attach them to their parent assemblies.

% Full systems
\paragraph{Full systems}
% Our scope is more limited, but our problems are larger
Previous work has proposed ``end-to-end'' multi-robot assembly planning and execution. For example, IKEABot is a multi-robot system for furniture assembly \citep{Knepper2013}. IKEABot takes a geometric assembly description as input, then synthesizes an assembly plan and coordinates the actions of delivery robots (which transport materials) and assembly robots (which attach parts to each other as prescribed by the assembly plan). \citet{Dogar2015} propose a system for multi-scale assembly with robot teams. The authors demonstrate their approach with an end-to-end hardware demo wherein a team of robots fastens a mock airplane wing panel to a mock wing box. 

More recent systems have continued to advance in scope and scale. \citet{Hartmann2020} presented a robust task and motion planning approach tailored to large-scale architectural assembly scenarios. Their approach tackled long-horizon planning problems by integrating task decomposition and motion planning for teams of mobile robots. \citet{Nagele2020} introduced Legobot, a framework specifically designed for coordinated multi-robot \LEGO{} construction, focusing on precise task assignment and collision-free trajectory scheduling.

While these recent systems have extended the complexity of assemblies that can be handled, they often emphasize fine manipulation tasks or detailed dynamic interactions between robots and assemblies---areas we intentionally abstract away. Our work, in contrast, targets larger-scale planning problems that involve coordinating many robots over expansive and hierarchically structured tasks. We specifically focus on geometric and spatial aspects such as assembly staging, team formation, and collision-free routing rather than fine-grained manipulation or physical dynamics. Our work complements these recent approaches by offering solutions aimed at scaling multi-robot coordination to even larger and more complex assemblies through a structured planning methodology that prioritizes geometric reasoning and efficient spatial and task allocation strategies.

% Assembly Planning 
\paragraph{Assembly planning}
Geometric assembly planning is the problem of determining trajectories along which components of an assembly can be brought into (or out of) mating position without interfering with the rest of the assembly. An assembly plan is often generated by first computing a disassembly plan (i.e., begin with a fully assembled model and plan how to remove each component) and then reversing the disassembly plan through time. \citeauthor{wilson1992geometric} introduced the concept of a ``non-directional blocking graph'' that encodes the geometric interactions/interferences between parts \citep{wilson1992geometric,Wilson1994}. This representation can be used to identify the directions in a part's configuration space in which it may be perturbed without interfering with the rest of the assembly. The non-directional blocking graph fits into the more general \emph{motion space} framework described by \citet{Halperin2000}. 

\Citet{Culbertson} used geometric considerations encoded by blocking graphs to derive partial ordering constraints specifically for multi-robot assembly planning thro\-ugh discrete optimization. Additionally, \citet{Rodriguez2020} proposed methods leveraging geometric pattern recognition and machine learning to efficiently generalize assembly sequences across varying product configurations. Our approach differs by assuming that a feasible geometric assembly plan has already been determined for each manufacturing project. Instead, we focus on orchestrating large-scale spatial and temporal coordination tasks such as staging and transporting assembly components rather than deriving initial geometric feasibility.

% Construction site layout
\paragraph{Construction site layout}
The geometric assembly plan specifies how to bring parts together. With large, complex assemblies, it is also important to determine where each subassembly will be constructed. This is closely related to the problem of facility layout planning, a well-studied topic in the literature \citep{Tompkins2010}. Discrete facility layout problems include the quadratic assignment problem \citep{Koopmans1957}. The design variables correspond to facility locations and the cost function is the sum of pairwise distance costs between facilities. Continuous facility layout (CFL) problems take the form of geometric packing problems (packing many small shapes into a larger shape) with similar pairwise distance costs \citep{Heragu1990}. A recent and comprehensive review by \citet{Perez-Gosende2021} surveys various methodologies and advancements in facility layout planning, covering both discrete and continuous approaches, and highlighting trends toward incorporating dynamic constraints and robotic navigation considerations. In our setting, the quality of a particular construction layout depends not only on the distance between related ``facilities'' (assembly construction areas) but also on the traversability of the inter-facility spaces through which robots must travel.

% Team Forming 
\paragraph{Team forming}
In scenarios where a team of robots must collaborate to transport a large payload, it is necessary to determine the ``carrying'' configuration of the robots relative to the payload. This problem is related to grasp planning in both single and multi-robot manipulation problems. Four cooperative manipulation protocols are proposed by \citet{Rus1995} for multi-robot planar manipulation of furniture. A distributed system for multi-robot collaborative transport is proposed by \citet{Fink2008}. A multi-robot grasp and \emph{regrasp} planner based on constraint satisfaction programming is proposed by \citet{Dogar2019}, for scenarios where a team of robots must work together to put together an assembly. \citet{Tariq2018} present a grasp coordination method for two-robot load sharing in collaborative transport tasks. They assume that the first agent's grasp has already been selected, and they select (from a finite number of candidate grasps) the second agent's grasp to optimize a load sharing objective function.

Recent approaches by \citet{Aswale2023} and \citet{Dai2025} have addressed coalition formation and scheduling problems in heterogeneous multi-robot systems. \citeauthor{Aswale2023} propose centralized algorithms to optimally or heuristically form coalitions of robots with varied skillsets, focusing on task scheduling under stochastic travel times. \citeauthor{Dai2025} present a decentralized reinforcement learning approach enabling dynamic team formation and task allocation, particularly emphasizing real-time responsiveness and adaptability in changing environments.

\citet{Ramchurn2010} study the problem of coalition formation with spatial and temporal constraints (CFSTP). Their setting involves tasks with deadlines and service durations and agents that can service these tasks by convening in teams at prescribed spatial locations. Different agents have different ``skills,'' which determine their effectiveness at servicing specific tasks. The size of a coalition required to service a given task depends on the skills of the team members. Though CFSTP involves deadlines and task service durations, it does not include intertask precedence constraints. We build upon the ideas presented by \citeauthor{Ramchurn2010} to include intertask precedence constraints and present algorithms that scale to large problems.

% Collision-free routing
\paragraph{Collision-free routing}
A very large body of work exists on distributed collision avoidance in continuous space. Two approaches relevant to our methods are artificial potential fields \citep{Khatib1985} and Reciprocal Velocity Obstacles (RVO) \citep{VanDenBerg2008}. Potential functions define vector fields that can be used to inform robots' continuous control signals. Repulsive potentials can push a robot away from obstacles and other robots, attractive potentials can draw a robot toward goals, rotational potentials can push a robot around obstacles or other robots, etc. Multiple potential functions can be composed to create control laws that simultaneously pursue multiple objectives. For example, \citet{Fink2008} use various potential field compositions with a finite state controller to enable distributed collaborative object transport.

RVO prevents collisions by placing constraints in neighboring robots' velocity spaces. It assumes that each robot's desired velocity is known to all other robots. In a convex environment where two robots are trying to reach different goals and we wish to minimize the sum of their travel times, RVO is an optimal collision avoidance strategy \citep{VanDenBerg2008}. Though this guarantee of optimality is not assured in scenarios with more than two simultaneously interacting robots, RVO leads to very efficient collision avoidance in sparse interaction settings. Though RVO is prone to gridlock in some scenarios, it can still work in settings with more dense interaction. We use these ideas as key components in our full-stack implementation to achieve a distributed, collision-free execution strategy.

\section{Environment}\label{sec:LEGO_environment}
We model the environment as a 3-dimensional Euclidean space. Robots, objects, and assemblies are modeled as rigid bodies. We assume a fleet of identical, cylinder-shaped transport robots. Objects and assemblies may have arbitrary $3$D geometry. The factory floor is a plane perpendicular to the vertical axis of the world coordinate frame $\worldframe.$ Robots are constrained to move only on the $2$D plane of the factory floor, and their orientation remains fixed (a robot's configuration is fully determined by its $x$- and $y$-position on the floor). The joint configuration space of the entire system (all robots and objects together) is the collision-free subset of the Cartesian product of their individual configuration spaces. 

\subsection{Assemblies}
An assembly consists of two or more components, whose prescribed configurations relative to the assembly frame are defined by transformation matrices. A component can be a single object or a subassembly with its own set of components. A project specification details one or more assemblies to be built, and may additionally group subsets of each assembly's components into an ordered sequence of build phases. A component may be added to its parent assembly only if the associated build phase is active. When all components in a build phase have been incorporated, the next build phase becomes active.

\begin{figure*}[tb]
  \centering
  \includegraphics[width=\textwidth]{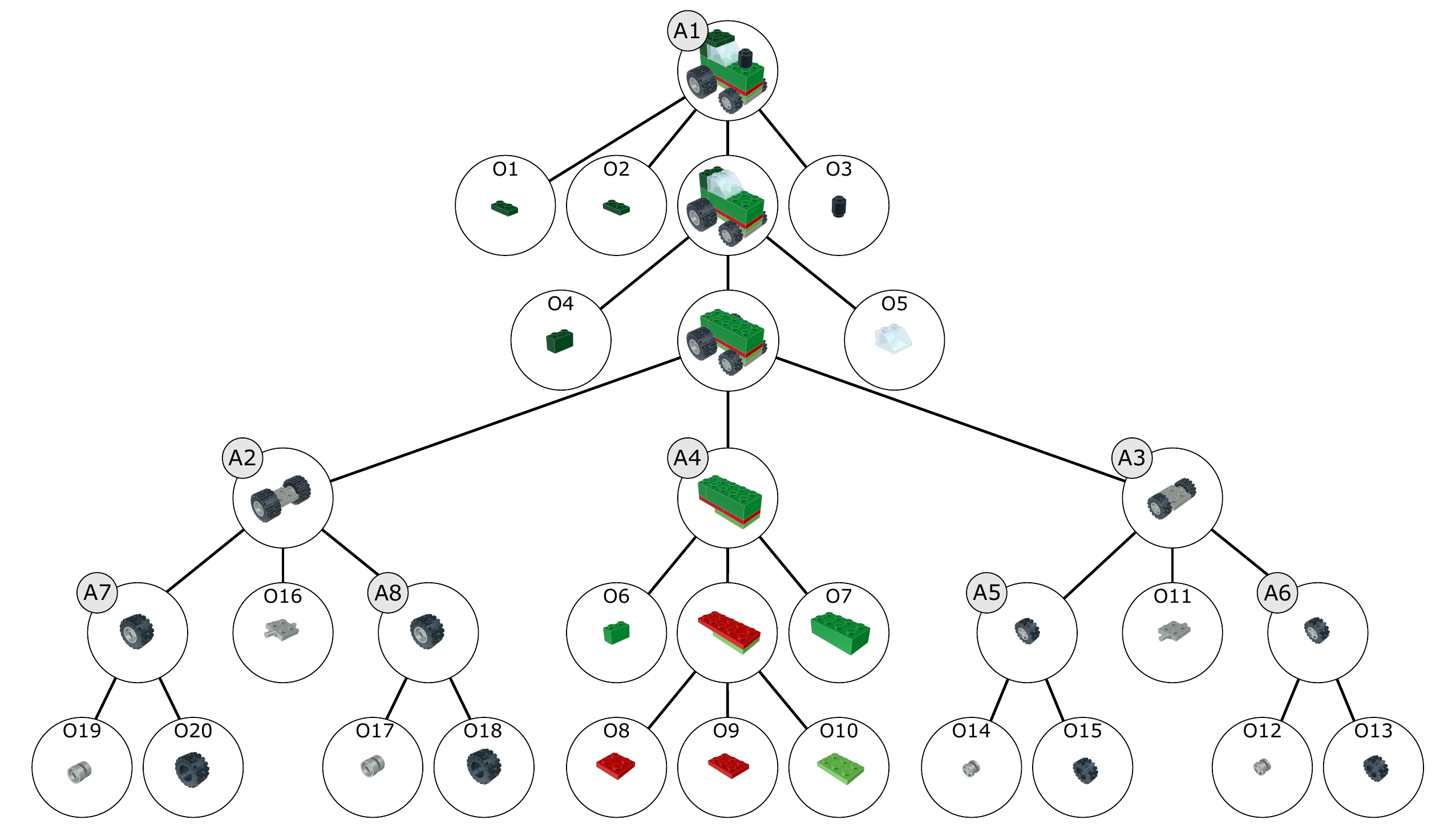}
  \caption{A visualization of the project specification for our example tractor assembly. The final assembly is composed of three subassemblies and a few individual parts.
 }
  \label{fig:tractor_model_spec}
\end{figure*}

In this work, we use \LEGO{} models to evaluate our algorithms. \LEGO{} models offer a convenient framework for defining large assemblies that are often composed of smaller assemblies in addition to individual parts. Throughout this work, we provide illustrations with assemblies that are defined using the \LDraw{}\footnote{\url{https://www.ldraw.org}} file specification, an open-source tool for describing \LEGO{} bricks and models. Some of the models used in our examples can be found in the LDraw Official Model Repository.\footnote{\url{https://omr.ldraw.org}} We designed others ourselves using LeoCAD,\footnote{\url{https://www.leocad.org}} an open-source CAD software tool for defining LDraw models.\footnote{This work is neither sponsored, authorized, nor endorsed by \LEGO{}, \LDraw{}, or LeoCAD.} A graphical depiction of the tractor assembly is shown in \cref{fig:tractor_model_spec}.

\subsection{Transport Units}
When a single robot or a team of robots transports an object or assembly, the robots and cargo together are referred to as a transport unit. To collect a payload, a robot (or team of robots) must move into carrying formation at a defined pickup location. Once at the pickup location, the cargo moves into its carrying configuration relative to the robots. When the cargo is secured in its carrying configuration, the transport unit may begin to move through the environment.

For a given object or assembly, the environment model employs a geometric heuristic to specify how many robots must participate in the transport unit. The planner must then identify an appropriate carrying formation for the team. Since the geometric ``team size'' heuristic is closely related to our method for determining transport unit formation, we introduce both in \cref{sec:configure_transport_units}.

The nominal configuration space of a transport unit is the same as that of a robot---i.e., the transport unit may translate along the floor of the environment, but may not rotate or move vertically. The velocity of the transport unit is constrained according to 
\begin{align}
  \Vert\posvel\Vert \leq \max (\vmaxrobot{} - \rectvolume{}\cdot\velvolumefactor{}, \vminrobot{}), \label{eqn:transport_unit_speed_limit}
\end{align}
where $\vmaxrobot{}$ denotes the maximum speed permitted for an unladen robot, $\rectvolume$ denotes the volume of the smallest hyperrectangle that completely encloses the transport unit, $\velvolumefactor{}$ is a scaling parameter, and $\vminrobot{}$ is a lower bound on the speed limit. This simple heuristic speed limit law is a proxy for a more sophisticated model that might account for robot and cargo dynamical properties, actuator constraints, and other considerations. The speed limit rule adds a layer of realism and complexity to the problem of collision-free navigation, as moving entities differ both in size and in the speed at which they can travel.

When a transport unit reaches the delivery location for its cargo, the payload is moved into a prescribed staging configuration. The dropoff procedure may not begin until the associated build phase of the cargo's parent assembly is active. After the cargo is moved into its staging configuration, the transport unit disbands, allowing the robots to break from carrying formation and attend to other tasks. Meanwhile, the assembly component is moved from its staging configuration into its final configuration relative to the parent assembly frame. Upon reaching the goal configuration, the component is ``captured'' and locked into place as part of the parent assembly.

\subsection{Methods Overview}\label{sec:LEGO_methods}
In this work, we start with a project specification that details the geometry of the parts and where they are located within the assembly. A graphical representation of a project specification is shown in \cref{fig:tractor_model_spec}. To fulfill a project specification, our autonomous multi-agent robotic assembly system creates and executes a construction plan. An overview of the major components of this process is provided in \cref{fig:methods_overview}. The plan is created in three primary stages and then executed with a distributed collision avoidance strategy.

\paragraph{Configure transport units (\cref{sec:configure_transport_units})}Our system first determines how many robots will be needed and where each robot will be positioned relative to the payload in order to transport each object and assembly. A few of the transport unit configurations for the tractor project are visualized in \cref{fig:transport_unit_configs}.
  
\paragraph{Construct staging plan (\cref{sec:construct_staging_plan})}The system determines where to build each assembly and where each component of that assembly will be deposited. Each assembly is constructed in its own staging area---a circular region on the factory floor. We determine the component dropoff locations by minimizing the distance from the dropoff location within the staging area to the final configuration of the component within the parent assembly. We attempt to arrange the staging areas so that every assembly can be transported in a straight line from its staging area to the staging area of its parent assembly without entering any other staging areas. An example staging plan for the tractor project is shown in \cref{fig:tractor_staging_plan}.

\paragraph{Allocate transport tasks (\cref{sec:collaborative_task_allocation})}We construct a new type of operating schedule that incorporates collaborative transport tasks and discrete build phases. Partial schedules for the tractor project are shown in \cref{fig:partial_schedule,fig:sub_graph_with_dummy_nodes}. To allocate tasks to individual robots and robot teams, we build upon the task allocation algorithm described by \citet{Brown2020a}.

\paragraph{Collision avoidance (\cref{sec:rvo_route_planning})}To execute the construction plan, the robots must perform their assigned delivery tasks while avoiding collision with each other. We propose a distributed online strategy where each agent follows a reactive velocity control policy consisting of a switching controller, a dispersion component, and a collision-avoidance controller.

\begin{algorithm*}[htbp!]
  \small
    \caption{Greedy Carrying Position Optimization}\label{alg:greedy_carrying_positions}
    \begin{algorithmic}[1]
      \State \textbf{Input:}
          \State \quad $c$: Vertices of the convex hull of the payload footprint (candidate robot carrying positions)
          \State \quad $n$: The number of robots required for carrying the payload (computed as described in \cref{sec:configure_transport_units})
      \State \textbf{Output:}
          \State \quad A set of $n$ robot positions selected from $c$
      \vspace{0.5em}
      \Function{select\_carry\_positions}{$c$,$n$}
        \If{$n = \vert v \vert$}
          \State \Return $c$
        \EndIf
        % \State $\bestidxs \gets [1, \ldots, n]$
        \State $\bestidxs \gets$ $n$ indices drawn uniformly from $1{:}\vert v \vert$ without replacement
        \State $\updatedflag = \textsc{true}$
        \While{$\updatedflag$}
          \State $\updatedflag \gets \textsc{false}$
          \For{$\candidateidxs \in \idxneighbors{\bestidxs}$} 
            \If{$\idxscore{c[\candidateidxs]} > \idxscore{c[\bestidxs]}$}
              \State $\bestidxs \gets \candidateidxs$
              \State $\updatedflag \gets \textsc{true}$
            \EndIf
          \EndFor
        \EndWhile
        \State \Return $c[\bestidxs]$
      \EndFunction
      \Function{score}{$pts$}
        \State $c_1 \gets \min_{i \in 1{:}\vert pts \vert}(\Vert pts[i]-pts[i+1] \Vert)$ \label{eqn:min_neighbor_dist}
        \State $c_2 \gets \sum_{i \in 1{:}\vert pts \vert} \Vert pts[i]-pts[i+1] \Vert$ \label{eqn:sum_neighbor_dist}
        \State $c_3 \gets \min_{p_1 \in pts, p_2 \in pts}(\Vert p_1-p_2 \Vert)$
        \State \Return $c_1 + \frac{0.5}{\vert pts \vert} c_2 + \frac{0.1}{\vert pts \vert^2} c_3$ \label{eqn:min_inner_dist}
      \EndFunction
      \Function{neighbors}{$\candidateidxs$}
        % \State \Return $\{\candidateidxs^\prime \mid \forall i \in 1{:}\vert \candidateidxs \vert, \Vert \candidateidxs[i] - \candidateidxs^\prime[i] \Vert \leq 1, \forall (i,j) \in \candidateidxs^\prime, i \neq j\}$ %all unique indices whose elements are within $\pm1$ of the elements of $\candidateidxs$
        \State \Return $\{\candidateidxs^\prime \mid \Vert \candidateidxs[i] - \candidateidxs^\prime[j] \Vert \leq 1, \forall i,j \in 1{:}\vert \candidateidxs \vert, i \neq j\}$
      \EndFunction
    \end{algorithmic}
  \end{algorithm*}

\section{Configuring Transport Units}\label{sec:configure_transport_units}

\begin{figure*}[htbp!]
  \centering
  \includegraphics[width=0.75\textwidth]{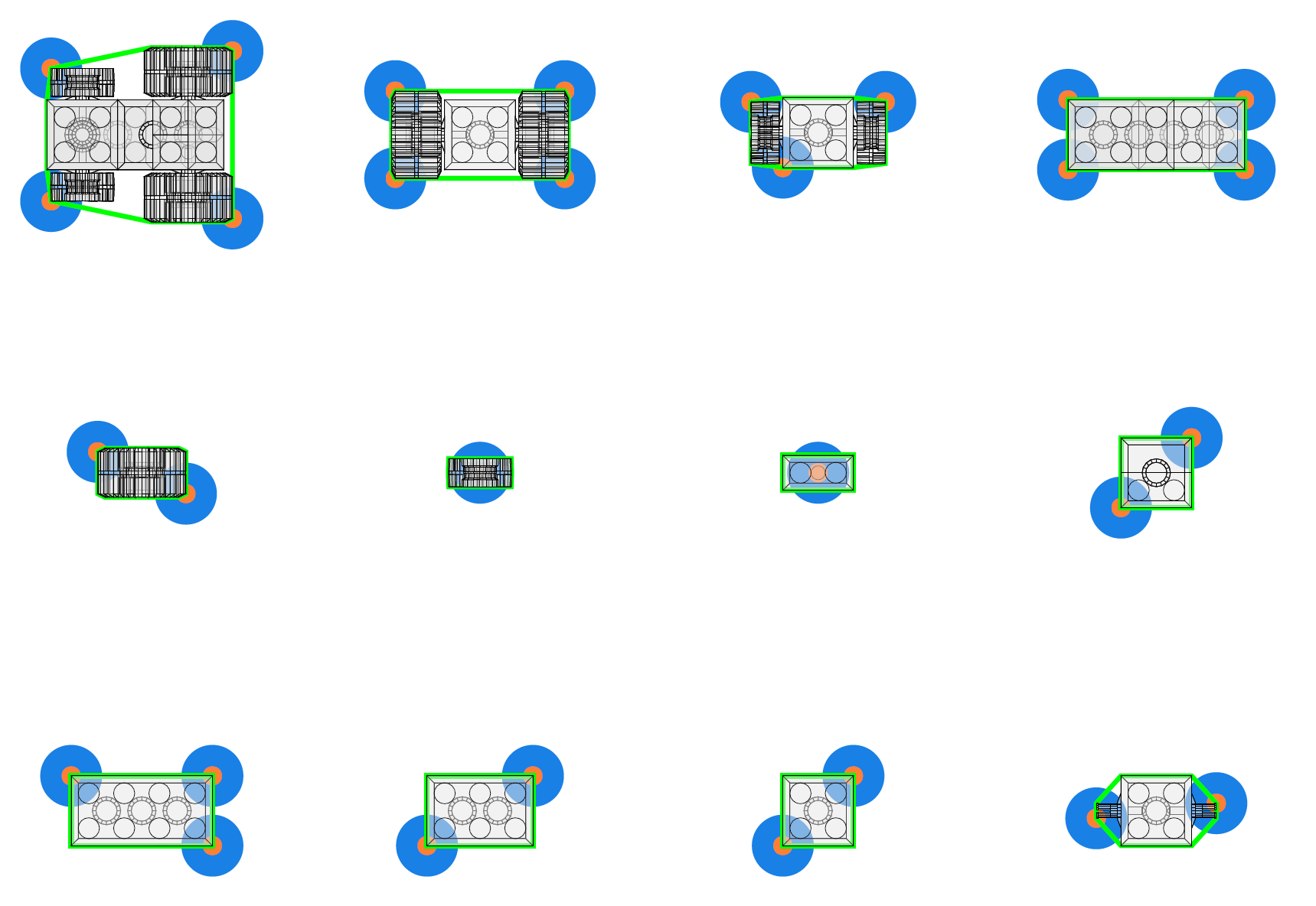}
  \caption{A visualization of the transport unit configurations for several of the assemblies and objects associated with the tractor project. The payload geometry is shown in black and white. The convex hull is highlighted in green. Robots are shown as blue disks, with their carrying positions highlighted with smaller orange disks.}
  \label{fig:transport_unit_configs}
\end{figure*}

For each payload to be transported, it is necessary to determine (a) how many robots should participate in the transport unit, and (b) where they should be positioned relative to each other and the payload. In a real-life collaborative transport scenario, these considerations would depend on many factors, including total mass and mass distribution of the payload, structural properties of the payload, ``grippability'' of the payload (i.e., where and how can robots securely grasp the payload), robot actuator limits, and the shape of both the payload and the robots.

Determining physically viable configurations for transport teams is a complex and multifaceted problem in its own right. Our investigation is strategically limited to the coordination aspect of multiple agents. Therefore, we adopt a heuristic geometric method that approximates spatial feasibility, allowing us to abstract away from the nuanced physical dynamics of real-world assembly. Our heuristic is primarily designed to ensure that payloads of various shapes and sizes require different numbers of robots, thus effectively exercising our planning algorithm. While inspired by practical considerations, the specific formulas were selected empirically to yield reasonable robot distributions rather than precise physical constraints. This approach enables us to focus on the algorithmic challenges of coordinating agent behaviors within spatial constraints, which aligns with the core contributions of our work. The remainder of this section describes our heuristic approach in more detail.

Given an object $o$, let $c$ represent the convex hull of the projection of $o$ onto the $x$-$y$ plane---i.e., the ``footprint'' of $o$ on the factory floor. We assume that $o$ has no curved surfaces, or, alternatively, that its geometry has been approximated such that there are no curved surfaces and that the error between true and approximated geometry is very small compared to the size of a robot. Because we assume no curved surfaces in the geometry of $o$, $c$ is a polygon. We define the set of candidate ``support points'' as the vertices of $c$. These are the locations at which robots may be placed to support the object in a transport unit.
The reason for limiting the candidate carrying positions to the vertices of the convex hull is that the object may have arbitrary non-convex shape; there could, for example, be a gaping hole in the middle of the object, such that a robot placed underneath the hole would not be able to carry any weight. If a robot positions itself at a vertex of the footprint, we can guarantee that it will be directly beneath a solid part of the object.

Let $l$ represent the length of $c$ defined as the maximum distance between any two points in $c$, and let $w$ represent the width defined as the maximum distance between any two points in $c$ projected onto the plane normal to the direction in which $l$ is measured (the values of $l$ and $w$ are the first and second singular values of the matrix whose columns are formed from the coordinates of the vertices of $c$). Let $p$ denote the perimeter of polygon $c$, and let $\robotradius$ denote the robot radius. Finally, let $N$ denote the number of edges in $c$ whose lengths are less than $2\robotradius$ (we cannot place two robots at both ends of any such edge). We wish to find $n$, the number of robots required to transport $o$.

Note that $\minrobots = ( p / (\pi \robotradius) ) \backslash 1$ is a lower bound on the number of disks of radius $\robotradius$ that can fit around the perimeter of $c$, where the backslash operator  $\cdot \backslash  \cdot$ denotes integer division (i.e., $\cdot \backslash 1$ denotes rounding down to the nearest integer). 
The value of $n$ is determined by
\begin{numcases}{n = }
  \scriptstyle \max(1,\;\min(\vert c \vert - N,\; \min( \minrobots,\; 2 \sqrt{\minrobots})) \backslash 1), & $\text{if} \ w \geq 2\robotradius$ \label{eqn:num_robots_case_one} \\
  \scriptstyle \max (1,\; \min (\minrobots,\; 2) ),\;  & \text{otherwise,}
\end{numcases}
where $\vert c \vert$ denotes the number of vertices of $c$. 
The case defined by \cref{eqn:num_robots_case_one} applies for payloads where the width of the payload is greater than twice the robot radius. It maximizes the number of robots under the constraint that there must be enough feasible carrying positions $\vert c \vert - N$ and the number of robots must not exceed $\minrobots$ or $2\sqrt{\minrobots}$. This second term is used to ensure that the number of robots grows sublinearly with increasing footprint perimeter for large payloads.
The second case ensures that long skinny objects are carried by just two robots. 

If $n=1$, the single robot carrying position is directly below the center of the minimum-radius hypersphere that fully encloses the payload. If $n>1$, the carrying positions are selected from the candidate positions according to the greedy hill climbing optimization procedure outlined in \cref{alg:greedy_carrying_positions}. The idea is to initialize a vector of indices into the points of $c$, and then iteratively improve those indices by trying all neighboring indices whose elements are within $\pm1$ of the elements of the corresponding current indices. The score of a given set of carrying positions is a linear combination of three terms: the first term (\cref{eqn:min_neighbor_dist}) measures the minimum distance between consecutive points; the second term (\cref{eqn:sum_neighbor_dist}) measures the sum of these neighbor-neighbor distances; the third term (\cref{eqn:min_inner_dist}) measured the minium distance between any two points in the set. This particular score function is a hand-engineered heuristic to encourage carrying configurations where the robots are as spread out as possible. Examples of generated transport unit configurations for the tractor assembly, subassemblies, and objects are shown in \cref{fig:transport_unit_configs}.

\section{Hierarchical Geometry Approximation}\label{sec:hierarchical_geom}

\begin{figure}[htbp!]
  \centering
  \includegraphics[width=0.5\textwidth]{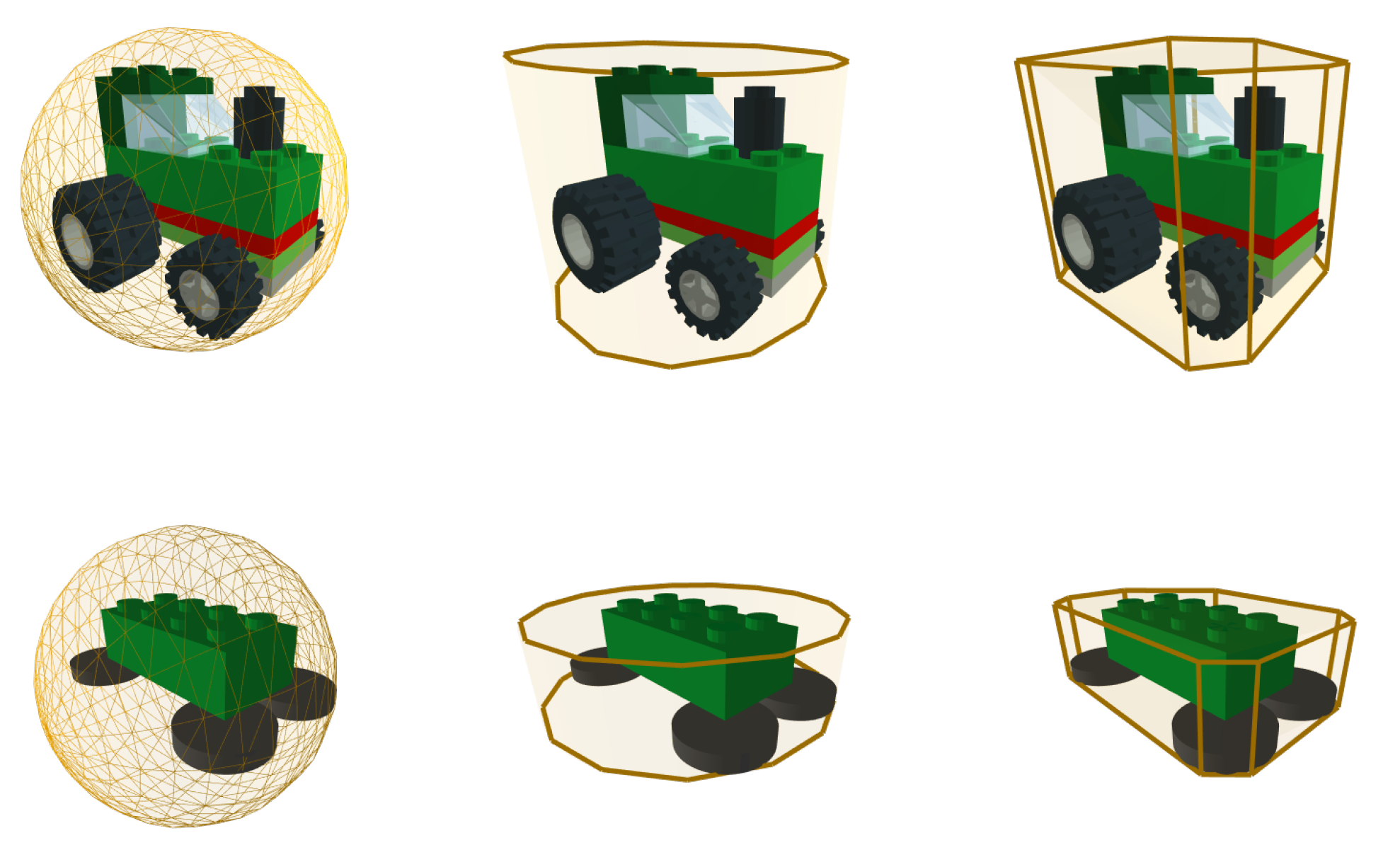}
  \caption{Over-approximated geometries (sphere, vertical cylinder, octagonal prism) for the final tractor assembly (top row) and for a transport unit.}\label{fig:overapprox_demos}
\end{figure}

When creating the construction plan, we often need to reason about the distance between geometric sets representing assemblies, components, and robots. In our construction planning framework, \emph{base geometry} refers to the original and detailed geometric representations of the assemblies, components, and robots involved in the manufacturing project. These geometries are typically complex and non-convex, reflecting the true physical shapes and sizes of the objects. Accurately computing distances between these intricate geometries can be computationally intensive, which motivates our use of convex overapproximations of the base geometry.

We use three types of bounding geometries (spheres, vertical cylinders, and vertical octagonal prisms; see \cref{fig:overapprox_demos}) to efficiently reason about geometries at various planning stages, including staging (\cref{sec:construct_staging_plan}) and collision avoidance (\cref{sec:rvo_route_planning}). Specifically, bounding spheres are employed primarily for tasks that benefit from rapid distance checks or simpler geometries, such as deconfliction during route planning. Vertical cylinders are used to determine staging locations, and vertical octagonal prisms offer tighter approximations, providing more precise vertical and volumetric information useful for tasks requiring detailed geometry, including calculating heuristic-based constraints (e.g., maximum transport speeds) and conducting accurate simulations.

The bounding geometry for each individual object is computed directly from the base geometry. For assemblies, the bounding geometry can be computed either from the exact geometry of the components or from their approximations. Although the latter is more efficient and thus preferable for larger assemblies, it results in looser-fitting approximations. For all assemblies discussed in this paper, we opt for the former, more precise approach. Additionally, we apply the same approach to compute bounding geometries for each transport unit based on the geometry of the associated payload and robot team.

To compute a bounding sphere, we solve a quadratic program to find the point that minimizes the maximum $L_2$ distance to any point in the input set. This can be done more efficiently with algorithms like the one proposed by \citet{Larsson2008}. To compute a bounding vertical cylinder, we solve the same quadratic program used for computing a bounding sphere, but with the input set projected onto the horizontal plane. We then compute the top and bottom of the cylinder by finding the maximum and minimum values of the input set projected onto the vertical axis. This same method is used to determine the top, bottom, and sides of the bounding vertical octagonal prism, but we additionally impose constraints so that each vertical face of the prism has a width of at least some minimum positive value. This ensures that each prism will have all eight of its sides.

\section{Constructing a Staging Plan}\label{sec:construct_staging_plan}
The global staging plan defines the staging location of each assembly---that is, where on the factory floor the assembly will be built. For each assembly, the assembly subplan consists of a sequence of build-phase subplans. A build phase subplan prescribes the dropoff location for each component in a given build phase relative to the staging location of its parent assembly.

We begin by constructing each build phase subplan independently. The idea is to select the dropoff locations so as to minimize the distance between each component's dropoff location and its final configuration in the assembly, subject to the constraints that (a) the transport units will not overlap with each other if they simultaneously occupy their prescribed dropoff zones, and (b) the transport units will not overlap with the partial assembly. We also want to ensure that no transport unit will have to wait for another to move before it can access its dropoff location. We select the dropoff locations by solving a convex radial layout optimization problem of the form
\begin{align}
  \qquad \qquad & \underset{{\CompAngle{1:n+1}}}{\text{minimize}} &  & \sum_{i = 1}^{n} (\CompAngle{i} - \DesiredCompAngle{i})^2 \label{eqn:ring_opt_obj} \\
  & \text{subject to} 
    & & \CompAngle{i+1} - \CompAngle{i} \geq \HalfWidth{i} + \HalfWidth{i+1}, \ i \in 1{:}n \label{eqn:ring_opt_overlap} \\
  & & & 0 \leq \CompAngle{i} \leq 2\pi, \ i \in 1{:}n \\
  & & & \CompAngle{n+1} - \CompAngle{1} = 2\pi, \ i \in 1{:}n, \label{eqn:ring_opt_dummy}
\end{align}%
where the decision variables $\CompAngle{1:n}$ denote the angular coordinates of the $n$ components' dropoff locations relative to the assembly, $\DesiredCompAngle{i}$ represents the angular coordinate of the goal configuration of component $i$ in the assembly, $\CompRadius{i}$ denotes the radius of component $i$'s bounding cylinder, and $\HalfWidth{i}=\arcsin(\CompRadius{i}/(\CompRadius{i}+\hat{\AssemblyRadius{}}))$ is the radial ``half width'' of component $i$ (half the width of component $i$'s ``slice of the pie''), where $\hat{\AssemblyRadius{}}$ represents the radius of the assembly bounding cylinder. The $n$ components are first sorted in order of increasing $\DesiredCompAngle{}$, so the non-overlap constraints around the rim of the assembly cylinder can be encoded in the convex form of \cref{eqn:ring_opt_overlap}.
The constraint in \cref{eqn:ring_opt_dummy} is a ``wrap around'' constraint that uses the dummy variable $\CompRadius{n+1}$ to apply the non-overlap constraint between component $n$ and component $1$. 

A radial layout problem is shown in \cref{fig:demo_ring_opt}. In \cref{fig:demo_ring_opt}, the black circle represents the current bounding cylinder of the parent assembly, and the blue circles represent the bounding cylinders of individual components positioned around it during the current build step. The radius of the black circle is determined by the maximum extent of the existing assembly geometry at that build stage, while each blue circle radius corresponds to the size of the respective component being added.

% 1 = angle
% 2 = radius
% 3 = base_radius
% 4 = extend
% 5 = angle label
% 6 = label
% 7 = arc label
% 8 l top label offset
\newcommand{\circwithconstraints}[8]{%
  \pgfmathsetmacro\ra{((#2+#3)*(1pt/1cm)} % radius of circle center
  \pgfmathsetmacro\rb{#4*(1pt/1cm)} % radius of guide line
  \pgfmathsetmacro\thetac{asin(#2/(#2+#3))}
  % \node[] () {\thetac};
  \draw (#1:#5-#8) -- (#1:\rb); % center line
  \node[comp,minimum size=2*#2,fill=white] () at (#1:\ra) {#6};
  \draw[blue,->] (0:#5) arc (0:#1:#5) node[pos=#7,circle,fill=white]{$\theta_{#6}$};
  \draw (#1-\thetac:#3) -- (#1-\thetac:#4); % guide line right
  \draw (#1+\thetac:#3) -- (#1+\thetac:#4); % guide line right
  \draw[red,<->] (#1:#4-#8) arc (#1:#1-\thetac:#4-#8) node[midway,circle,fill=white]{$\Delta_{#6}$};
  \draw[red,<->] (#1:#4-#8) arc (#1:#1+\thetac:#4-#8) node[midway,circle,fill=white]{$\Delta_{#6}$};
}

% 1 = angle
% 2 = label radius
% 3 = length
% 4 = label position
% 5 = index
\newcommand{\desiredangle}[5]{
  \draw[red,dashed] (0,0) -- (#1:#3);
  \draw[red,->] (0:#2) arc (0:#1:#2) node[circle,pos=#4,fill=white]{$\hat{\theta}_{#5}$};
}
\begin{figure}[tb]
  \centering
  \begin{tikzpicture}[
    scale=1.0,
    every node/.style={inner sep=0.5pt,outer sep=0pt},
    boundingcircle/.style={circle,outer sep=0pt,inner sep=0pt},
    assembly/.style={boundingcircle,draw,thick},
    comp/.style={boundingcircle,draw=blue},
  ]
    \draw (0,0) -- (0:4.25cm);

    \desiredangle{321}{0.8cm}{1.8cm}{1.05}{5}
    \desiredangle{301}{0.8cm}{1.8cm}{0.83}{4}
    \desiredangle{200}{0.8cm}{1.8cm}{0.82}{3}
    \desiredangle{130}{0.8cm}{1.8cm}{0.65}{2}
    \desiredangle{45 }{0.8cm}{1.8cm}{0.50}{1}

    \circwithconstraints{333    }{1.00cm}{1.75cm}{4.25cm}{1.35cm}{5}{1.03}{0.2cm}
    \circwithconstraints{287    }{1.25cm}{1.75cm}{4.25cm}{1.35cm}{4}{0.83}{0.2cm}
    \circwithconstraints{200    }{0.75cm}{1.75cm}{4.25cm}{1.35cm}{3}{0.82}{0.2cm}
    \circwithconstraints{130    }{1.10cm}{1.75cm}{4.25cm}{1.35cm}{2}{0.65}{0.2cm}
    \circwithconstraints{45     }{1.00cm}{1.75cm}{4.25cm}{1.35cm}{1}{0.50}{0.2cm}

    \node[assembly,minimum size=3.5cm] (a) at (0,0) {};

  \end{tikzpicture}
  \caption{An example solution to a radial layout optimization problem. Circles $1$, $2$ and $3$ are placed precisely at their respective desired orientations relative to the center circle. Circles $4$ and $5$, however, are forced to split the difference because they would overlap if placed at their desired orientations.}
  \label{fig:demo_ring_opt}
\end{figure}
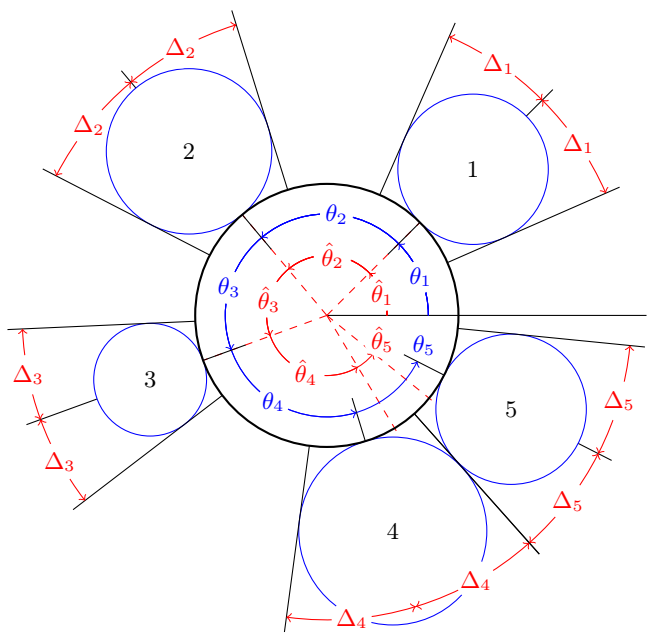

The bounding cylinder of an assembly is dynamic; it may expand with each build step as more components merge into the structure. For each radial layout optimization iteration, the initial bounding cylinder corresponds to the assembly's state before that particular build step. A build phase staging area is established for each build phase subplan, and it's designed to be the smallest possible cylinder that encompasses: (1) the current state of the assembly's bounding cylinder. (2) all designated drop-off zones, and (3) the staging area from the preceding build step (for the assembly's inaugural build phase, this previous staging area does not exist).

When the bounding cylinder of the assembly does not provide enough circumference to accommodate all components, a multi-tiered optimization process becomes essential. We accomplish this by using component prioritization and solving the radial layout problem iteratively. Components are ranked based on their build step sequence and their respective radii, $\CompRadius{}$. Early-stage build components are given spatial priority (positioned closer to the assembly). The maximum feasible set of components that can surround the current bounding cylinder is determined. A radial layout optimization then occurs for this subset of components. Once placed, we recompute the bounding cylinder to include these new drop-off zones and iterate until all components are placed. This layered structuring efficiently uses the factory floor, especially as the assembly and component counts increase.

Following the formulation of all build phase subplans, it is time to position each assembly in relation to its parent. Every leaf assembly (assemblies that lack child assemblies) has its staging area earmarked as its construction zone. Assemblies are organized in a hierarchical tree structure, where each parent assembly is composed of one or more child assemblies or individual objects. For instance, in the example tractor assembly depicted in \cref{fig:tractor_model_spec}, assemblies $7$ and $8$ are child assemblies of assembly $2$, while assemblies $2$, $3$, and $4$ are child assemblies of assembly $1$. Assemblies without child assemblies (e.g., assemblies $7$ and $8$) are called leaf assemblies, as they appear at the end of the assembly tree. We then move up the assembly tree (from initial objects to final assemblies), defining the construction zone of each parent assembly by solving a radial layout optimization problem. In these layout problems, the target angle $\DesiredCompAngle{}$ of each child is defined by the angle of the dropoff zone of that child relative to its parent assembly.

\begin{figure*}[htb!]
  \centering
  \begin{subfigure}{0.49\textwidth}
    \centering
    \includegraphics[width=\textwidth]{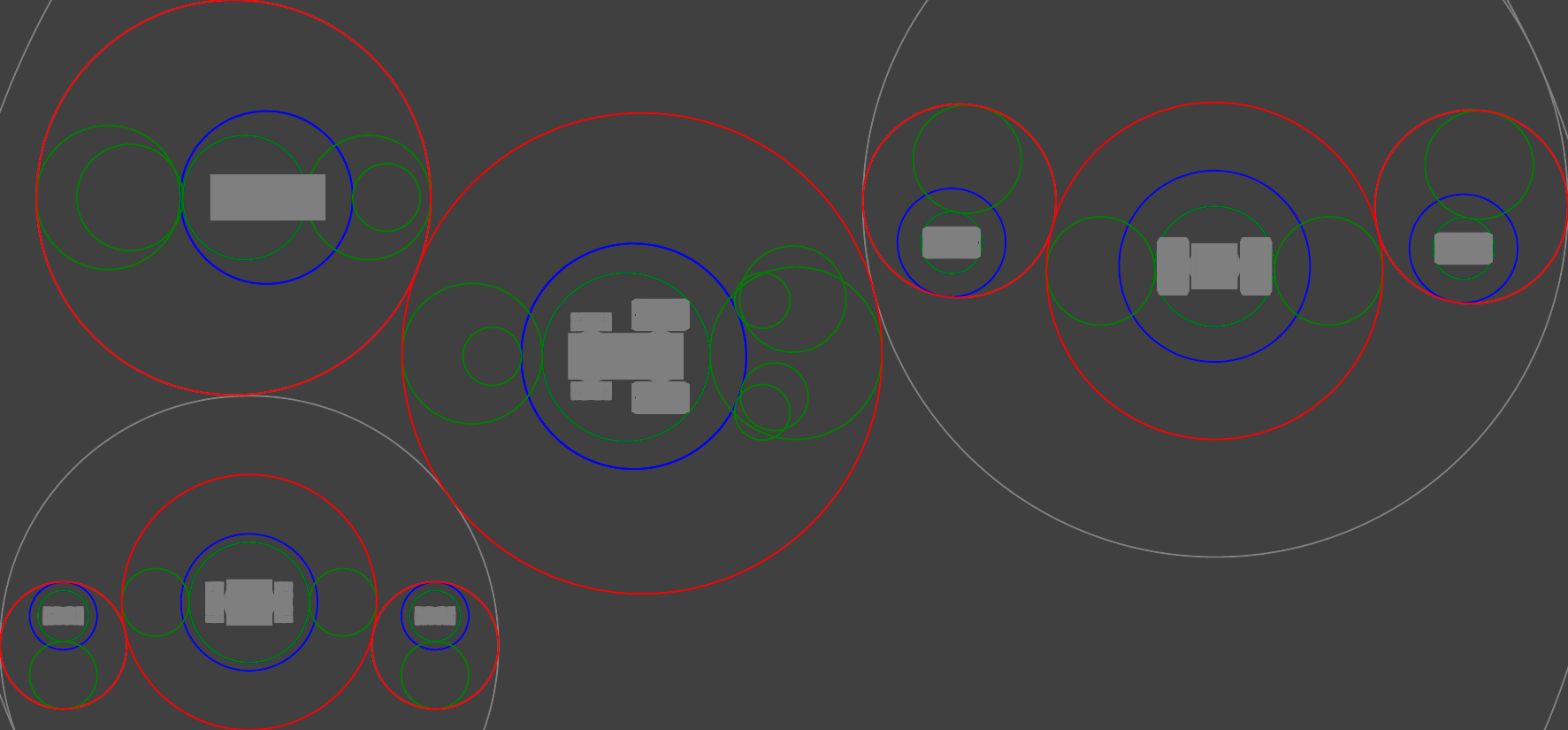}
    \caption{Tractor staging plan with no buffer ($20$ parts, $8$ assemblies).}
    \label{subfig:tractor_no_buffer}
  \end{subfigure}%
  \hfill
  \begin{subfigure}{0.49\textwidth}
    \centering
    \includegraphics[width=\textwidth]{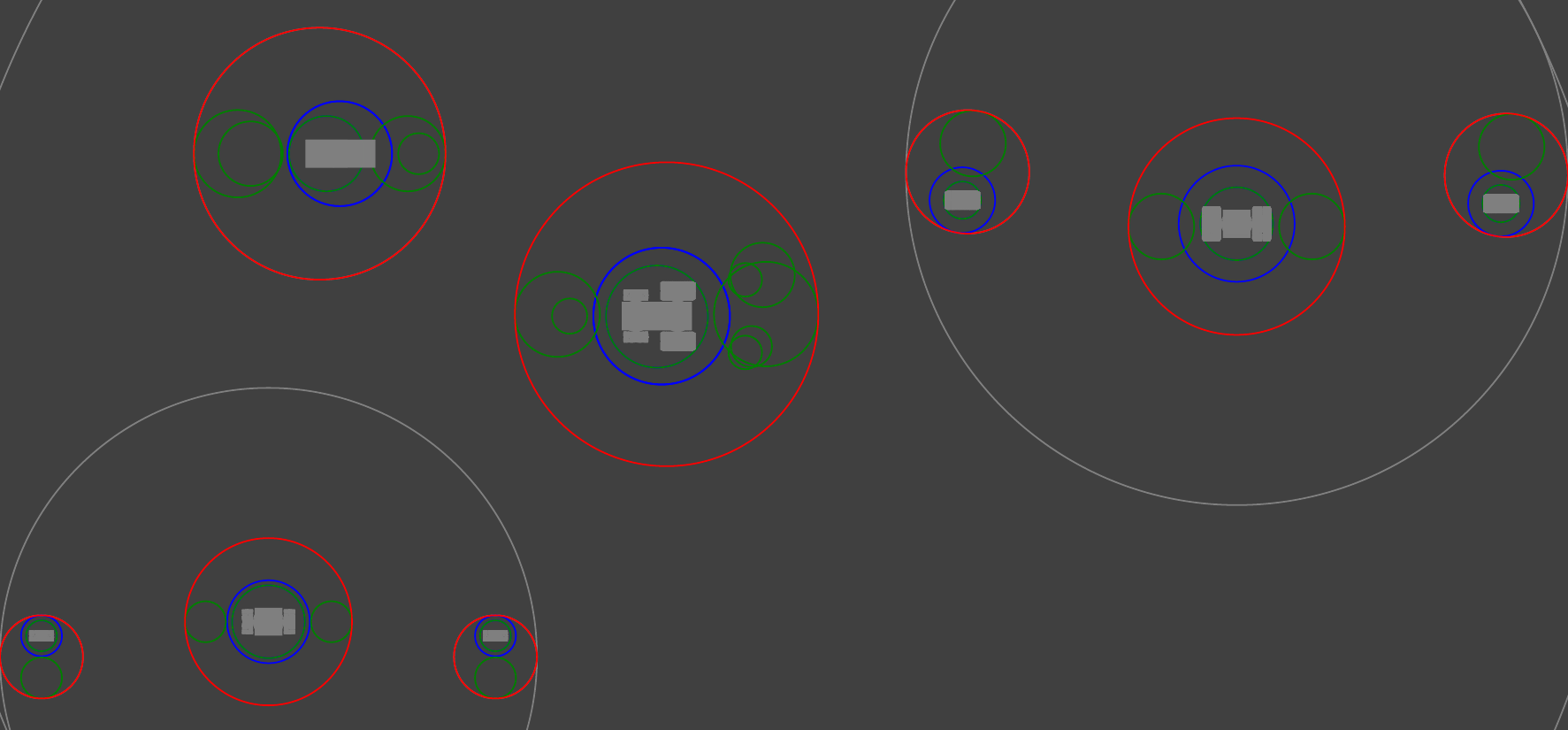}
    \caption{Tractor staging plan with buffer ($20$ parts, $8$ assemblies).}
    \label{subfig:tractor_with_buffer}
  \end{subfigure}

  \vspace{1em}

  \begin{subfigure}{0.49\textwidth}
    \centering
    \includegraphics[width=\textwidth]{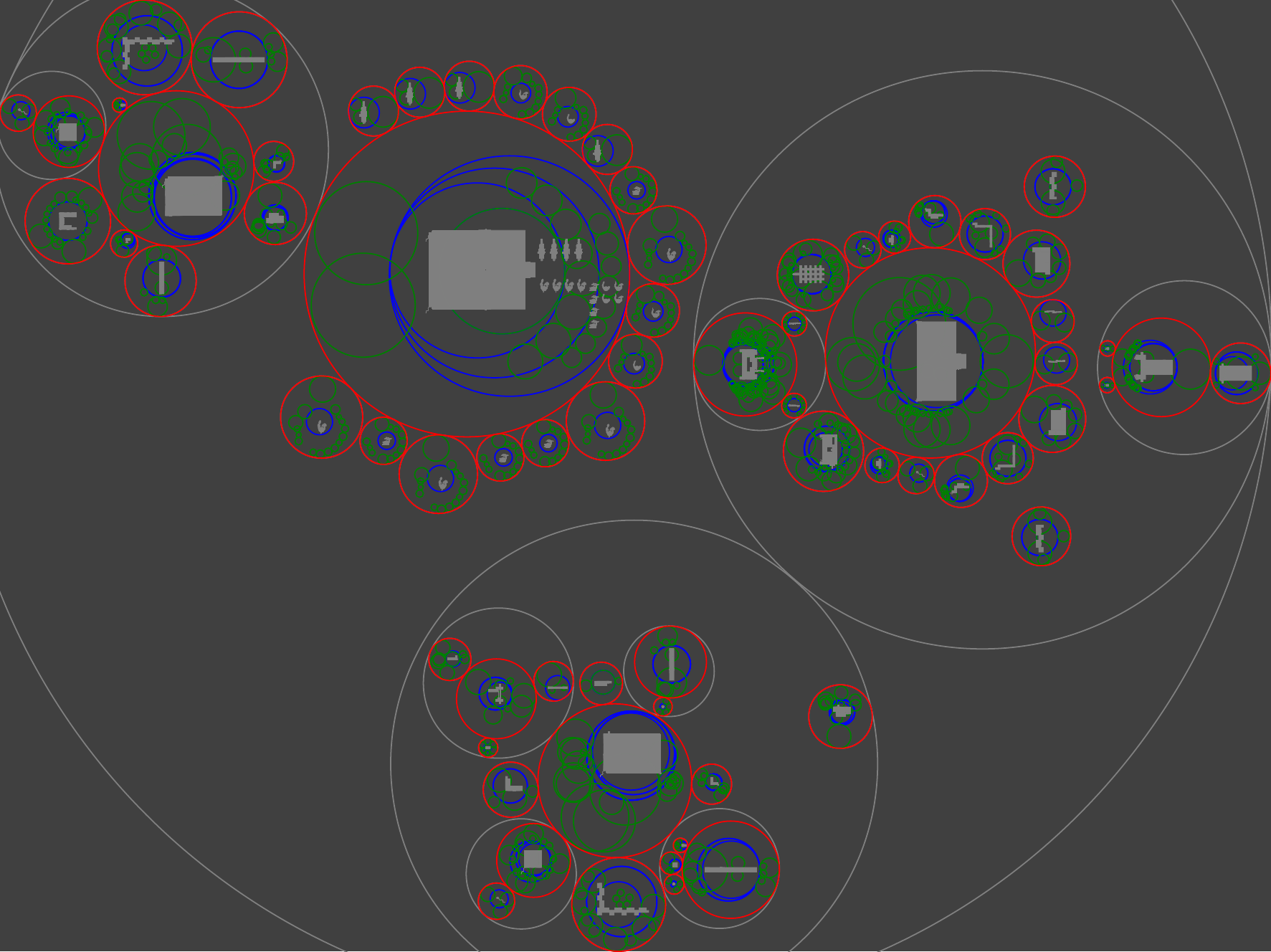}
    \caption{King's Castle staging plan with no buffer ($761$ parts, $70$ assemblies).}
    \label{subfig:kings_castle_no_buffer}
  \end{subfigure}%
  \hfill
  \begin{subfigure}{0.49\textwidth}
    \centering
    \includegraphics[width=\textwidth]{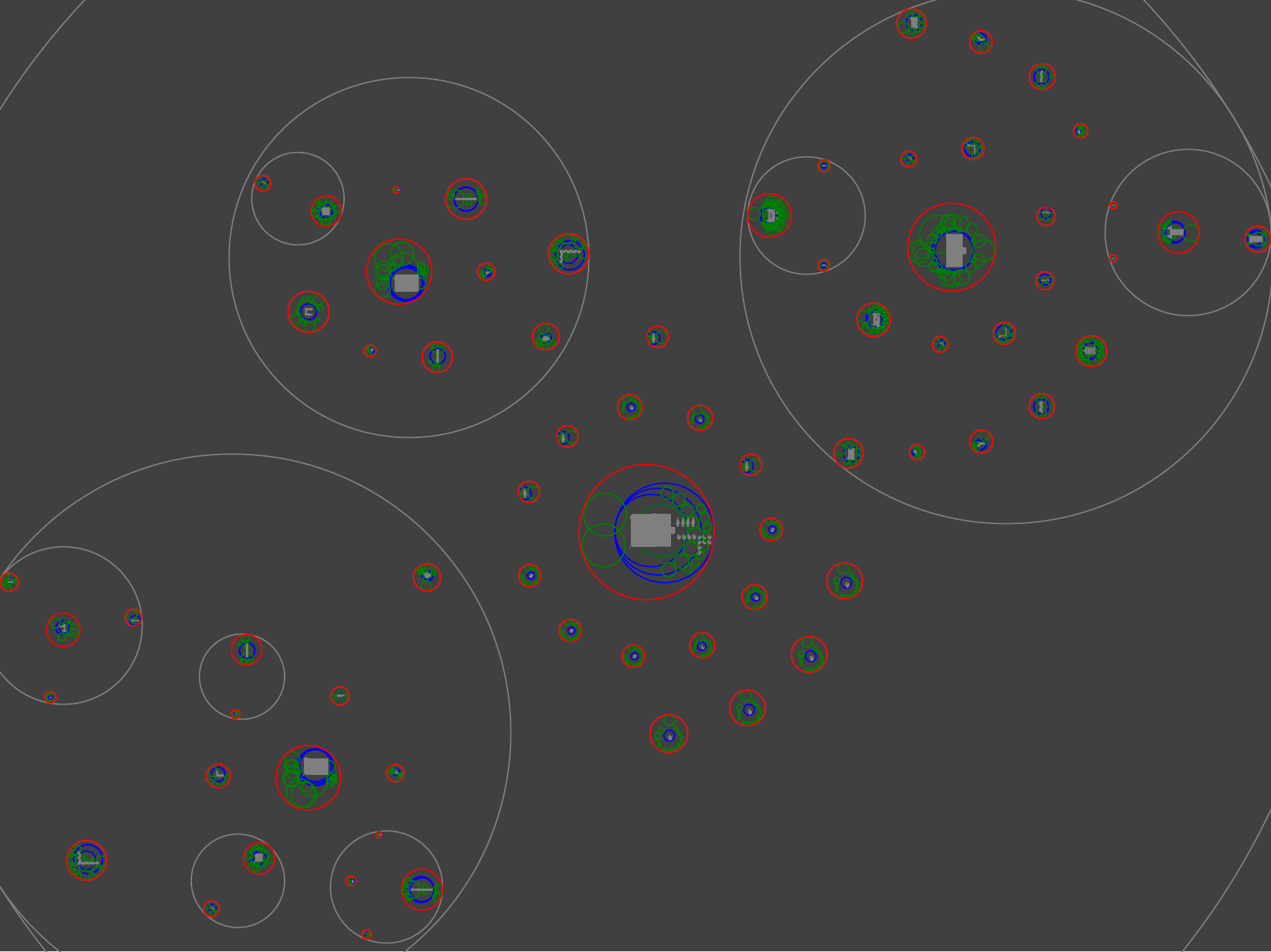}
    \caption{King's Castle staging plan with buffer ($761$ parts, $70$ assemblies).}
    \label{subfig:kings_castle_with_buffer}    
  \end{subfigure}%

  \caption{Staging plans for the Tractor and King's Castle projects. Red circles represent the staging areas of the assemblies, blue circles are the final bounding cylinders of the assemblies, and green circles represent the dropoff zones for assembly components. Gray circles are added to show how the nested staging areas fit around each other.}
  \label{fig:tractor_staging_plan}
\end{figure*}

\Cref{fig:tractor_staging_plan} shows staging plans for the tractor project and for the King's Castle project (discussed in more detail in \cref{sec:LEGO_demos}). \Cref{subfig:tractor_no_buffer} and \cref{subfig:kings_castle_no_buffer} are constructed precisely as described above, while \cref{subfig:tractor_with_buffer} and \cref{subfig:kings_castle_with_buffer} include an additional buffer radius around each construction zone. During our simulations, we found that adding a buffer helps avoid workspace crowding and reduces the burden on collision avoidance logic. King's Castle, composed of $70$ assemblies and $761$ parts, illustrates how layered concentric rings can efficiently use factory space compared to expanding a single ring to fit all assemblies. Although dropoff zones may appear to overlap in the figure, this is merely an artifact of visualizing all zones simultaneously, irrespective of assembly stage; these zones are, in fact, temporally deconflicted.

Introducing a buffer radius, however, does increase the distances robots must travel, potentially extending overall construction time. Selecting an optimal buffer size thus involves a trade-off between collision avoidance ease and transportation efficiency. While employing more advanced routing algorithms that dynamically consider time-dependent availability of assembly zones could mitigate buffer sizes and maintain efficiency, exploring these approaches is beyond the scope of the current work. Nevertheless, our radial layout optimization is computationally efficient (see \cref{tab:demo_results}) and sufficiently flexible to quickly adapt buffer sizes based on specific operational priorities.

\section{Team Forming and Task Allocation}\label{sec:collaborative_task_allocation}
% New kind of operating schedule
With a global staging plan that defines where all assemblies will be built, collected, and transported, the next task is to determine which specific robots will be involved in the transport of each object or assembly. 

\subsection{The Operating Schedule}
An operating schedule $\ProjectSchedule = (\ScheduleVertices,\ScheduleEdges)$ is a directed acyclic graph (DAG) where each vertex $\vtx \in \ScheduleVertices$ corresponds to a discrete high-level event or activity and an edge $(\vtx \rightarrow \vtxTWO) \in \ScheduleEdges$ denotes a precedence constraint, requiring that the activity associated with $\vtx$ must be completed before the activity associated with $\vtxTWO$ may begin \citep{Brown2020a}. In our setting, the operating schedule includes the following node types:

{
  \tikzexternaldisable
\begin{itemize}
  \item \textbf{$\ObjectStart{}$} %
  \ObjectStartN \;defines the initial state of an object.
  \item \textbf{$\RobotStart{}$} %
  \RobotStartN \;defines the initial state of a robot.
  \item \textbf{$\RobotGo{}$} %
  \RobotGoN \;defines a navigation task for a single robot from one location to another.
  \item \textbf{$\AssemblyStart{}$} %
  \AssemblyStartN \;is a checkpoint node that must be passed before work may begin on an assembly.
  \item \textbf{$\OpenBuildStep{}$} %
  \OpenBuildStepN \;is a checkpoint at which the referenced build step becomes active. This checkpoint is reached for the initial build step of an assembly as soon as the \AssemblyStart{} checkpoint is passed. For each subsequent build step, the \OpenBuildStep{} checkpoint is reached as soon as the previous build step has been completed.
  \item \textbf{$\FormTransportUnit{}$} %
  \FormTransportUnitN \;defines the task of loading a payload onto a team of robots in formation. This task may only begin when the robots are in carrying formation. During the \FormTransportUnit{} task, the robots remain in place as the payload is lowered into its carrying configuration.
  \item \textbf{$\TransportUnitGo{}$} %
  \TransportUnitGoN \;defines the task of transporting a payload to its dropoff zone. During transport the robots and payload remain in rigid formation.
  \item \textbf{$\DepositCargo{}$} %
  \DepositCargoN \;defines the task of unloading a payload from a transport unit. The robots remain in formation until the payload has been lifted into its staging configuration, at which time the transport unit disbands and the robots are free to break from formation and attend to other tasks.
  \item \textbf{$\LiftIntoPlace{}$} %
  \LiftIntoPlaceN \;defines the task of moving an assembly component from its staging configuration to its target configuration in the assembly. This task is accomplished without participation of any robots. We assume that a manipulator robot is available to move the component from its staging configuration to its final configuration.
  \item \textbf{$\CloseBuildStep{}$} %
  \CloseBuildStepN \;is a checkpoint at which the build step is completed. This checkpoint is reached once all \LiftIntoPlace{} tasks associated with the referenced build step have been completed. 
  \item \textbf{$\AssemblyComplete{}$} %
  \AssemblyCompleteN \;is a checkpoint that marks an assembly as complete, meaning that it is ready to be collected by a transport unit.
  \item \textbf{$\ProjectComplete{}$} %
  \ProjectCompleteN \;is a checkpoint marking the project as complete.
\end{itemize}
}

\subsection{Task allocation and Team Forming as a Graph Repair Problem}
In the original task assignment formulation of \citet{Brown2020a}, a single-robot-per-task structure is hard-coded into the \milp{} constraints. In that work, \citeauthor{Brown2020a} encoded the decision variable as a binary assignment matrix $A \in \textbf{B}^{(n+m) \times m}$, where $n$ is the number of robots, $m$ is the number of assignments, and $A_{ij}=1$ indicates that robot $i$ is assigned to transport object $j$. The first $n$ rows of $A$ corresponded to real robots while the last $m$ rows corresponded to the robots after they performed a previous assignment. For example, if $A_{ij}=1$ and $A_{j+n, k}=1$ then robot $i$ is assigned to deliver object $j$ and then assigned to deliver object $k$. Hence this formulation cannot be directly applied to our setting where robots are frequently required to work together as part of a transport unit.

\begin{figure*}[htbp!]
  \centering
  \includegraphics[width=\textwidth]{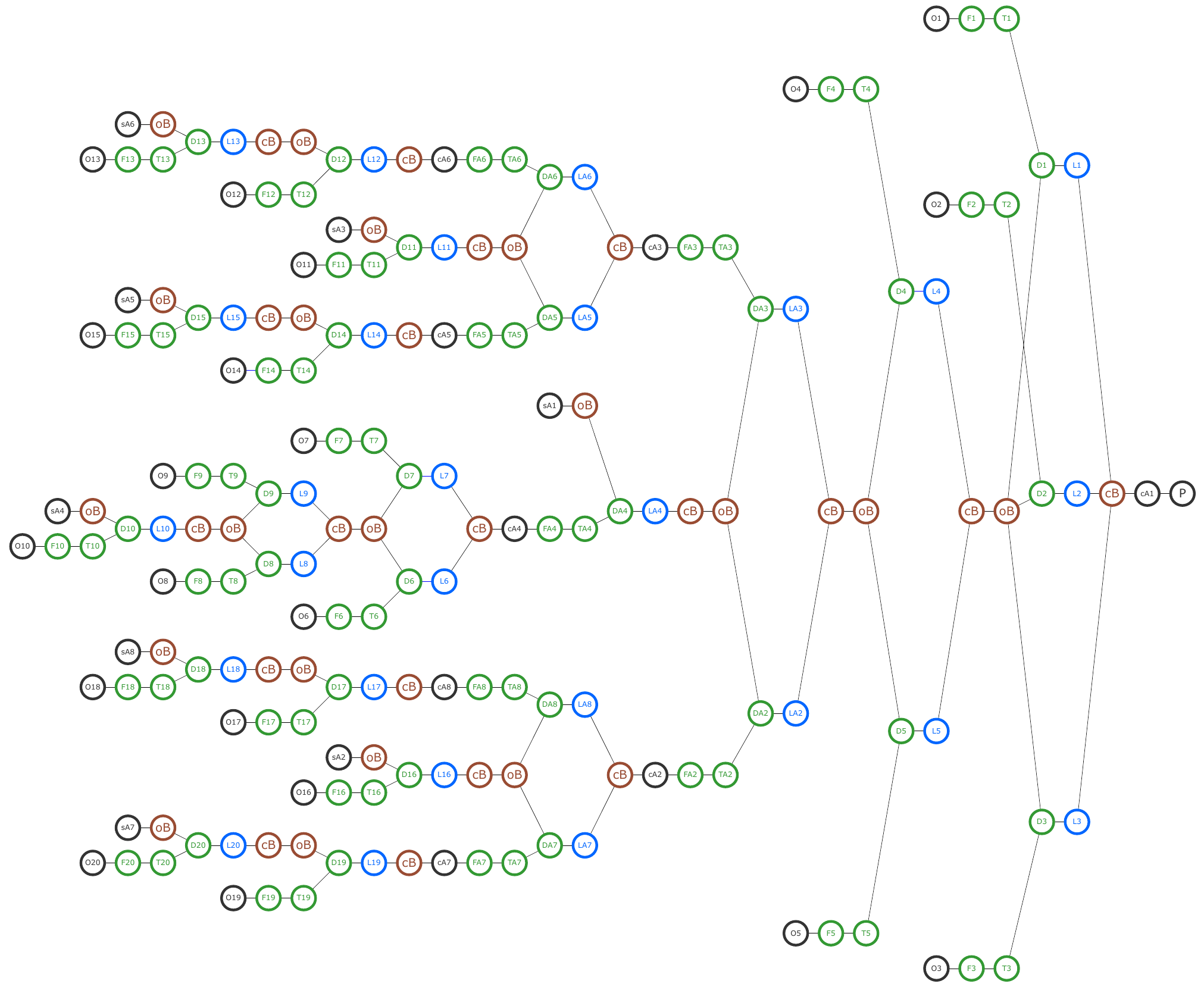}
  \caption{The partial schedule Tractor operating schedule. This schedule encodes all tasks that need to be performed and the precedence constraints between them. The partial schedule does not yet encode any assignments of robots to tasks. \RobotStart{} and \RobotGo{} nodes are hidden to emphasize the structure of the transport tasks.}
  \label{fig:partial_schedule}
\end{figure*}

\begin{figure*}[htbp!]
  \centering
  \includegraphics[width=0.8\textwidth]{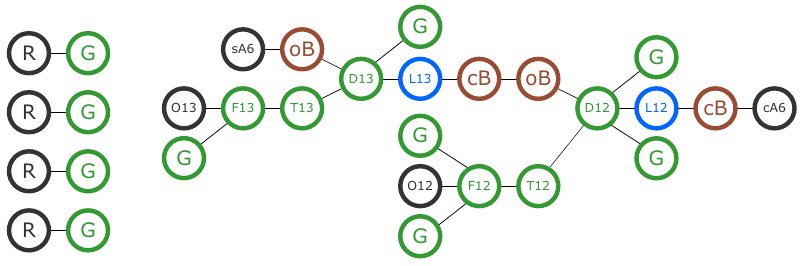}
  \caption{Assembly $6$ subgraph of the partial Tractor operating schedule. Nodes associated with a particular object or assembly are annotated with the ID of that object/assembly. Several free \RobotStart-\RobotGo{} pairs are also shown, emphasizing that tasks have not yet been allocated to specific robots.}
  \label{fig:sub_graph_with_dummy_nodes}
\end{figure*}

We introduce a more generic task assignment \milp{} formulation that makes it straightforward to deal with arbitrary project schedule structures---including those that incorporate collaborative transport tasks with varying numbers and configurations of robot teams. Our new formulation can be thought of as a graph repair problem: an initial graph is specified, but some required edges (in our setting, the assignment edges) are missing from the graph. A solver must determine where to add edges so as to satisfy the problem constraints and minimize some performance objective (in our setting, the makespan or, for multi-head projects, the sum of makespans).

Instead of solving for an assignment matrix, we solve for the adjacency matrix $\adjacencyMatrix{}{}$ of the project schedule, which directly encodes the edges of the project schedule. That is, the adjacency matrix $\adjacencyMatrix{}{}$ is the symmetric $|\ScheduleVertices| \times |\ScheduleVertices|$ matrix encoding of the adjacency relationships in our schedule graph
\begin{equation}
    \adjacencyMatrix{i}{j} =  
    \begin{cases}
        1 & \text{if } (i \rightarrow j) \in \ScheduleEdges \\
        0              & \text{otherwise}.
    \end{cases}
\end{equation}%

We first specify an initial schedule graph by adding edges corresponding to the various transport tasks. The structure of this initial schedule encodes the partial ordering of all tasks that need to be accomplished. However, the initial schedule is missing the assignment edges, which encode the assignments of tasks to robots. The sets of allowable edges to and from a given node are encoded by the helper functions $\EligiblePredecessors$ and $\EligibleSuccessors$, respectively. The sets of required edges to and from a given node are defined by the functions $\RequiredPredecessors$ and $\RequiredSuccessors$, respectively. The suffixes $\textsc{-Pred}$ and $\textsc{-Succ}$ are short for predecessors and successors.

\begin{table}[tb!]
  \centering
  \ra{1.2}
  \caption{Required predecessors and successors for schedule node types. The asterisk denotes that the number of predecessors/successors for a given type can vary between instances of that node type.}
  \begin{tabular}{@{}lll@{}}
  \toprule
  \multirow{2}{*}{Node Type} & \multicolumn{2}{c}{Eligible $/$ Required} \\
  \cmidrule{2-3}
  & Predecessors & Successors \\
  \midrule
  \ProjectCompleteN & \AssemblyCompleteN & \\
  \ObjectStartN & & \FormTransportUnitN  \\
  \AssemblyStartN & & \OpenBuildStepN  \\
  \AssemblyCompleteN & \CloseBuildStepN  & \FormTransportUnitN$/$\ProjectCompleteN  \\
  \OpenBuildStepN & \AssemblyStartN$/$\CloseBuildStepN & \DepositCargoN$^*$  \\
  \CloseBuildStepN & \LiftIntoPlaceN$^*$ & \AssemblyCompleteN$/$\OpenBuildStepN  \\
  \RobotStartN & & \RobotGoN  \\
  \RobotGoN & \DepositCargoN$/$\RobotStartN$/$\RobotGoN & \\
  \FormTransportUnitN & \ObjectStartN$/$\AssemblyCompleteN,\;\RobotGoN$^*$ & \TransportUnitGoN \\
  \TransportUnitGoN & \FormTransportUnitN & \DepositCargoN \\
  \DepositCargoN & \OpenBuildStepN,\;\TransportUnitGoN & \LiftIntoPlaceN,\;\RobotGoN$^*$ \\
  \LiftIntoPlaceN & \DepositCargoN & \CloseBuildStepN \\
  \bottomrule
  \end{tabular}
  \label{tab:required_neighbors}
\end{table}

A graph is valid if and only if the predecessors and successors of each node form supersets of the respective $\textsc{Required-}$ sets and subsets of the respective $\textsc{Eligible-}$ sets. In other words, each node must have at least the required number of edges to and from the right types of nodes, and no more than the allowable number of edges to and from the right types of nodes. The outputs of these four functions for each type of schedule node are shown in \cref{tab:required_neighbors}.

The initial schedule for the tractor project is shown in \cref{fig:partial_schedule}, with each node labeled by an index. Nodes referring to individual objects use numeric indices (e.g., L$14$ indicates \LiftIntoPlace{} for object $14$), while nodes referring to assemblies use the prefix A before the index (e.g., LA$4$ indicates \LiftIntoPlace{} for assembly $4$). This notation distinguishes actions on objects from those on assemblies throughout the schedule graph.

% adding dummy TransportUnitGo nodes.
Since some transport units have multi-robot teams, it is necessary to identify which role (i.e., which carrying position) a given robot is being assigned to. We provide this extra information by adding \RobotGo{} nodes. For each carrying position in a transport unit, one \RobotGo{} node is added as a predecessor to the associated \FormTransportUnit{} node, and one is added as a successor of the associated \DepositCargo{} node. Each placeholder \RobotGo{} node stores the destination or origin of its associated carrying position. These placeholder nodes are omitted from \cref{fig:partial_schedule} so as not to distract from the structure of the transport tasks. However, they are included in the visualization of the assembly $6$ subgraph in \cref{fig:sub_graph_with_dummy_nodes}. In this subgraph, there is a \RobotGo{} node as a predecessor of the \FormTransportUnit{} node for object $15$ and two \RobotGo{} nodes as predecessors for the \FormTransportUnit{} node for object $14$ (this object requires $2$ robots for transport). Similarly, \RobotGo{} nodes follow the $\DepositCargo{}$ nodes for those objects as well.

Given an initial schedule (e.g. \cref{fig:partial_schedule}) and the helper functions as defined by \cref{tab:required_neighbors}, the \milp{} formulation can be written as%
{\milpfontsize%
\begin{alignat}{5}
  & {\text{\milpobjectivefontsize minimize}} \span && \;\; & \sum \CompletionTime[\idxA], \quad \idxA \in \ProjectSchedule{}.\TerminalProjectNodes \label{eqn:milp_objective} \\[-0.6ex]
  & {\text{\milpobjectivefontsize subject to}} \span &&& \span \nonumber \\ % binary adjacency matrix
    & \quad \adjMtx{\idxA}{\idxB} = 1, \quad (\idxA \rightarrow \idxB) \in \ScheduleEdges \label{eqn:existing_edges} \eightspans \\[-0.6ex] % existing edges
    & \quad \adjMtx{\idxA}{\idxB} = 0, \quad (\idxA \rightarrow \idxB) \notin \EligibleEdges(\ProjectSchedule) \label{eqn:eligible_edges} \eightspans \\[-0.6ex] % eligible edges only
    & \quad \adjMtx{\idxA}{\idxB} = 0, \quad \idxA \in \ScheduleVertices , \quad \idxB \in \textsc{upstream}(\ProjectSchedule,\idxA) \label{eqn:downstream} \eightspans \\[-0.6ex] % no edges to upstream nodes
    & \quad \textstyle \sum_{\idxA \in \ScheduleVertices} \adjMtx{\idxA}{\idxB} \geq \ 
    \vert \RequiredPredecessors(\idxB) \vert, \quad \idxB \in \ScheduleVertices  \label{eqn:req_preds}  \eightspans \\[-0.3ex] % predecessor constraints
    & \quad \textstyle \sum_{\idxB \in \ScheduleVertices} \adjMtx{\idxA}{\idxB} \geq \ 
    \vert \RequiredSuccessors(\idxB) \vert, \quad \idxA \in \ScheduleVertices   \label{eqn:req_succs} \eightspans \\[-0.3ex] % successor constraints
    & \quad \textstyle \sum_{\idxA \in \ScheduleVertices} \adjMtx{\idxA}{\idxB} \leq \ 
    \vert \EligiblePredecessors(\idxB) \vert, \quad \idxB \in \ScheduleVertices   \label{eqn:eligible_preds} \eightspans \\[-0.3ex] % predecessor constraints
    & \quad \textstyle \sum_{\idxB \in \ScheduleVertices} \adjMtx{\idxA}{\idxB} \leq \ 
    \vert \EligibleSuccessors(\idxB) \vert, \quad \idxA \in \ScheduleVertices   \label{eqn:eligible_succs} \eightspans \\[-0.3ex] % successor constraints
    % & \quad \solutionCost \geq \CompletionTime[\idxA], \quad \idxA \in \ScheduleVertices  \eightspans \label{eqn:milp_objective} \\ % precedence constraints
    & \quad \CompletionTime[\idxA]\geq \StartTime[\idxA] + \processtime[\idxA], \quad \idxA \in \ScheduleVertices \label{eqn:duration} \eightspans \\ % duration constraints
    & \quad \StartTime[\idxB] - \CompletionTime[\idxA] \ \geq -M (1 - \adjMtx{\idxA}{\idxB}), 
    \ \idxA \in \ScheduleVertices , \ \idxB \in \ScheduleVertices  \eightspans \label{eqn:big_m_precedence} \\ % Big-M constraints with adjacency matrix
    & \quad \CompletionTime[\idxA] - (\StartTime[\idxA] + \processtime[\idxA](\idxB)) \ \geq -M (1 - \adjMtx{\idxA}{\idxB}), 
    \ \idxA, \idxB \in \ScheduleVertices \eightspans\label{eqn:big_m_duration}% Big-M constraints with adjacency matrix 
    \\
    & \quad \StartTime[] \in \reals_{+}^{\vert\ScheduleVertices\vert}, \quad % node start times
    \CompletionTime[] \in \reals_{+}^{\vert\ScheduleVertices\vert} , \quad % node end times
    \adjMtx{}{} \in \booleans^{\vert\ScheduleVertices\vert \times \vert\ScheduleVertices\vert} \eightspans % binary adjacency matrix
\end{alignat}}%%
where \cref{eqn:milp_objective} defines the sum-of-makespans objective, $\StartTime[]$ and $\CompletionTime[]$ encode the start and end times, respectively, for all vertices,
\cref{eqn:existing_edges} encodes all existing edges, 
\cref{eqn:eligible_edges} disqualifies ``illegal'' edges, and 
\cref{eqn:downstream} prevents any single edge from creating a cycle in the graph. 
It is still possible for multiple added edges to create a cycle, but this does not occur in solutions because it has infinite cost. \Cref{eqn:req_preds,eqn:req_succs} ensure that each vertex has at least the required number of incoming and outgoing edges, 
\cref{eqn:eligible_preds,eqn:eligible_succs} ensure that each vertex has no more than the maximum allowable number of incoming and outgoing edges, and 
\cref{eqn:duration} enforces the duration of each vertex.
\Cref{eqn:big_m_precedence,eqn:big_m_duration} encode ``big $M$'' inequality constraints that are activated/deactivate by the value of the associated binary variable---%
\cref{eqn:big_m_precedence} enforces precedence constraints between vertices only if there is an edge between them, and 
\cref{eqn:big_m_duration} encodes the duration of vertex $i$ if that vertex is updated by adding an edge from $\idxA$ to $\idxB$ ($\processtime[\idxA](\idxB)$ encodes the duration if the edge is added). This last constraint is necessary because the duration of a \RobotGo{} node depends on its destination (which is defined by the successor of the \RobotGo{} node), and durations are estimated based on distances between relevant locations and an assumed average robot speed. These estimated durations provide necessary inputs for defining timing constraints within the MILP formulation.

\subsection{Comparing Matrix Formulations}
% \subsubsection{Comparing ``Adjacency Matrix'' and ``Assignment Matrix'' Formulations}
If we consider the operating schedules that arise in the original \pctapf{} formulation of \citet{Brown2020a}, we recognize that the size of a schedule's adjacency matrix is greater than the size of the assignment matrix used to create the schedule. Hence, the number of discrete and continuous optimization variables is greater in an adjacency matrix \milp{} formulation than in a comparable assignment matrix \milp{} formulation. This prompts the question, ``how much does solver runtime increase with the adjacency matrix formulation compared to the original assignment matrix formulation?''
To quantify the slowdown that occurs when solving an ``assignment \milp{}'' vs. an ``adjacency \milp{}'', we evaluate three \milp{} variants:
\begin{itemize}
  \item \AssignmentMILP{} is the original task assignment \milp{} formulation proposed by \cite{Brown2020a}.
  \item \AdjacencyMILP{} is the new, generic \milp{} formulation described above.
  \item \SparseAdjacencyMILP{} implements the same \milp{} formulation as \AdjacencyMILP{}, but employs sparse variable containers and a pre-processing routine that instantiates optimization variables only for allowable edges. 
\end{itemize}

Since \AssignmentMILP{} is limited to single-robot-per-task settings, we compare these three \milp{} variants on the \pcta{} subproblems of the original \pctapf{} problem set used for the experiments in \cite{Brown2020a}. In our current problem setting, we require the added flexibility to allow for teaming of robots to transport objects. We evaluate the different approaches in a single-robot-per-task setting to demonstrate the computational cost associated with the increased flexibility.

\begin{figure*}[htb!]
  \centering
  \resizebox*{\textwidth}{!}{
  \begin{tikzpicture}[
    node distance = 0.1cm and 0.1cm,
    inner sep = 0pt,
    outer sep = 0pt,
    /pgfplots/label style = {font=\large},
    /pgfplots/legend style = {font=\normalsize},
    /pgfplots/every axis/.append style = {ymajorgrids},
    /pgfplots/ymax = 100,
    /pgfplots/ymin = 0.02,
  ]
    \node[] (a) {\input{r_AdjacencyMILP_RunTime.tex}};
    \node[above = of a] (a title) {\AdjacencyMILP{} Runtime (s)};
    \node[left = of a] (b) {\input{r_SparseAdjacencyMILP_RunTime.tex}};
    \node[above = of b] (b title) {\SparseAdjacencyMILP{} Runtime (s)};
  \end{tikzpicture}
  }
  \caption{Absolute runtime plotted for \SparseAdjacencyMILP{} (left) and \AdjacencyMILP{} (right).}
  \label{fig:adjacency_milp_runtimes}
\end{figure*}

\begin{figure*}[htb!]
  \centering
  \resizebox*{\textwidth}{!}{
  \begin{tikzpicture}[
    node distance = 0.1cm and 0.1cm,
    inner sep = 0pt,
    outer sep = 0pt,
    /pgfplots/label style = {font=\large},
    /pgfplots/legend style = {font=\normalsize},
    /pgfplots/every axis/.append style = {ymajorgrids},
    /pgfplots/ymax = 100,
    /pgfplots/ymin = 1.0,
  ]
    \node[] (a) {\input{r_AdjacencyMILP_runtimeratio.tex}};
    \node[above = of a] (a title) {\AdjacencyMILP{}/\AssignmentMILP{} Runtime Ratio};
    \node[left = of a] (b) {\input{r_SparseAdjacencyMILP_runtimeratio.tex}};
    \node[above = of b] (b title) {\SparseAdjacencyMILP{}/\AssignmentMILP{} Runtime Ratio};
  \end{tikzpicture}
  }
  \caption{Runtime ratio plotted for \SparseAdjacencyMILP{} (left) and \AdjacencyMILP{} (right) compared to \AssignmentMILP{}. Some results for high $\numtasks$, low $\numrobots$ categories are not very meaningful because both \AssignmentMILP{} and the \AdjacencyMILP{} variant reached the time limit.}
  \label{fig:adjacency_milp_comparison}
\end{figure*}

\begin{figure*}[htb!]
  \centering
  \resizebox*{\textwidth}{!}{
    \input{r_adjacency_group_histogram.tex}
  }
  \caption{Histograms summarizing the distributions of runtime ratios for \SparseAdjacencyMILP{}/\AssignmentMILP{} (left) and \AdjacencyMILP{}/\AssignmentMILP{} (right) aggregated over all problem instances.}
  \label{fig:adjacency_group_histogram}
\end{figure*}

The absolute runtimes for \AdjacencyMILP{} and \textsc{Spa\-rseAdjacencyMILP} are plotted in \cref{fig:adjacency_milp_runtimes}. \Cref{fig:adjacency_milp_comparison} shows how the distribution over runtime ratios for \textsc{AdjacencyMILP}/\AssignmentMILP{} and \textsc{SparseAdjacen\-cyMILP}/\AssignmentMILP{} vary between classess. \Cref{fig:adjacency_group_histogram} summarizes the runtime ratios of the two \AdjacencyMILP{} variants aggregated across all problem classes. Both variants are slower than \AssignmentMILP{} and the slowdown becomes more pronounced for $\numtasks \gg \numrobots$. However, \SparseAdjacencyMILP{} scales better than \AdjacencyMILP{}.

We have explored several pre-processing approaches to reduce the problem size (and hence, the solve time) of the adjacency matrix \milp{} formulation. \SparseAdjacencyMILP{} represents the most successful of those approaches. Though the increase in runtime compared to \AssignmentMILP{} is unfortunate, it is a necessary burden in exchange for the added flexibility to support robot teaming for transport tasks in our current problem set. We hope to explore alternative formulations and extensions of the \AssignmentMILP{} formulation to mutli-robot-per-task scenarios in the future.

% Modified greedy task assignment
% Prevents assignment to unavailable build steps
\subsection{Modified Greedy Task Allocation}\label{sec:modified_greedy_task_allocation}

{
  \newcommand{\TeamAssignment}{team\_assignment}
  \newcommand{\ft}{t_{\text{task}}}
  \newcommand{\OpenSteps}{active\_steps}
  \newcommand{\ActiveAssemblies}{active\_assemblies}
  \newcommand{\AvailableComponents}{available\_components}
  \newcommand{\cargo}{component}
  \newcommand{\AvailableRobots}{available\_robots}
  \newcommand{\BuildStep}{build\_step}
  \newcommand{\assignmentlist}{pairs}
  \newcommand{\goals}{goals}
\begin{algorithm*}[htb!]
\small
% \footnotesize
% \scriptsize
  \caption{Greedy assignment algorithm for collaborative transport tasks.}
  \label{alg:greedy_LEGO_assignment}
  \begin{algorithmic}[1]
  \State \textbf{Input:}
    \State \quad Operating schedule graph
    \State \quad Set of available robots, each with initial position and availability time
 \State \textbf{Output:}
    \State \quad Updated operating schedule graph with added edges encoding robot assignments
  \vspace{0.5em}
  \Procedure{\greedyLEGO}{}
      \State $\ActiveAssemblies \gets $ all assemblies in project
      \State $\AvailableRobots \gets $ all robots
      \State $\AvailableComponents \gets $ all raw materials
      \While{$\ActiveAssemblies$ is not empty}
        \State $\TeamAssignment \gets nothing$
        \State $t_{\min} \gets \infty$
        \State $target \gets nothing$
        \ForEach{$assembly \in \ActiveAssemblies$}
          \ForEach{$\cargo \in assembly.active\_step.unassigned\_components$}
            \If{$\cargo \in \AvailableComponents$}
              \State $\goals \gets \cargo.pickup\_positions$
              \State $\assignmentlist \gets \emptyset$
              \State $\ft \gets 0$
              \While{$\goals$ is not empty}
                \State $(robot,goal), t \gets \textsc{EarliestArrival}(\AvailableRobots,goals)$
                \State $\ft \gets \max(\ft,t)$
                \If{$\ft \geq t_{\min}$}
                  \State $\textsc{break}$
                \EndIf
                \State $\assignmentlist \gets \assignmentlist \cup (robot,goal)$
                \State $\AvailableRobots \gets \AvailableRobots \setminus \{robot\}$
                \State $\goals \gets \goals \setminus \{goal\}$
              \EndWhile
              \For{$(robot,goal) \in \assignmentlist$}
                \State $\AvailableRobots \gets \AvailableRobots \cup \{robot\}$
              \EndFor
              \If{$\ft < t_{\min}$}
                \State $\TeamAssignment \gets (\cargo,\assignmentlist)$
                \State $t_{\min} \gets \ft$
                \State $target \gets assembly$
              \EndIf
            \EndIf
          \EndFor
        \EndFor
        \State Add $\TeamAssignment$ to operating schedule
        \If{all component transport tasks for $target.active\_step$ are assigned}
          \If{$target.active\_step = target.terminal\_step$ }
            \State $\ActiveAssemblies \gets \ActiveAssemblies \setminus \{target\}$
            \State $\AvailableComponents \gets \AvailableComponents \cup \{target\}$
          \Else
            \State $target.active\_step \gets $ next build step
          \EndIf
        \EndIf
      \EndWhile
  \EndProcedure
  \end{algorithmic}
\end{algorithm*}
}

We demonstrated that the \milp{} solver struggles when $\numtasks \gg \numrobots$. This effect becomes even more pronounced for \SparseAdjacencyMILP{} than for \AssignmentMILP{}. For very large assemblies, optimal task assignment is intractable. This does not necessarily mean that the \milp{} solver cannot be used for large assemblies. On the contrary, it is frequently the case that the \milp{} solver identifies multiple feasible, though not necessarily optimal, solutions in its search for a certifiably optimal solution. If at least one such solution has been found before the time or iteration limit is reached, the solver will return the best feasible solution found so far along with an upper bound on the optimality gap.

Nevertheless, we also wish to have a suboptimal task assignment algorithm with runtime guarantees. To this end, we propose a greedy precedence-constrained coalition formation (\greedyLEGO) algorithm (\cref{alg:greedy_LEGO_assignment}) that accounts for collaborative transport tasks and the precedence constraints associated with build phases and nested subassemblies. This algorithm produces a suboptimal, yet feasible solution. In scenarios were time allows, we can then use this feasible solution to warm-start our optimization process for the \SparseAdjacencyMILP{} problem.

\greedyLEGO{} adds assignments for an entire transport unit (which may include multiple agents) at each iteration by a greedy selection of the transport unit-task pair. The transport unit is selected via the \textsc{EarliestArrival} subroutine of \greedyLEGO{} (\cref{alg:earliest_arrival_subroutine}), which is similar to the earliest completion first algorithm proposed by \citet{Ramchurn2010} for multi-agent coalition forming with spatial and temporal constraints. 

For each available transport task, a candidate transport unit is selected by greedily assigning robots to the designated carrying formation positions. The transport unit's lower bound pickup time is the maximum over the candidate robot team of the time required for each robot to reach its assigned pickup configuration. The transport unit and associated robot team with the lowest pickup time are added to the schedule. 

% In addition, \greedyLEGO{} limits the set of available tasks at a given assignment iteration to the tasks that belong to active build steps (i.e., build steps whose preceding build steps have all been fully assigned). % The addition of collaborative tasks and build phases requires us to redefine what constitutes an available task and an available robot.

\begin{algorithm}[htb!]
  \small
  \caption{Earliest Arrival Subroutine called by \greedyLEGO{}.}
  \label{alg:earliest_arrival_subroutine}
  \begin{algorithmic}[1]
    \State \textbf{Input:}
      \State \quad $robots$: Currently available robots with locations and availability times
      \State \quad $goals$: Goal locations to be assigned to robots
    \State \textbf{Output:}
      \State \quad $assignment$: Robot-goal assignment with the earliest arrival time
      \State \quad $t_{\min}$: The corresponding arrival time
    \vspace{0.5em}
    \Function{EarliestArrival}{robots,goals}
      \State $assignment \gets nothing$
      \State $t_{\min} \gets \infty$
      \ForEach{$robot \in robots$}
        \ForEach{$goal \in goals$}
          \State $t \gets $ earliest time at which $robot$ can reach $goal$
          \If{$t < t_{\min}$}
              \State $assignment \gets (robot,goal)$
              \State $t_{\min} \gets t$
          \EndIf
        \EndFor
      \EndFor
      \State \Return $assignment, t_{\min}$
    \EndFunction
  \end{algorithmic}
\end{algorithm}

In the original \pctapf{} setting, an unassigned task was considered available if all of its predecessors had been assigned. In our setting, availability of a task additionally requires that its build step be active. We require this modification because each \DepositCargo{} node is preceded by both a \TransportUnitGo{} node and an \OpenBuildStep{} node. Hence, the \DepositCargo{} task may not begin until its associated build phase becomes active. Therefore, if \greedy{} were to prematurely assign all robots to ``downstream'' build phases, it would essentially consign the whole fleet to wait with their cargo indefinitely (because no robots would be available to attend to the previous build phases). Hence, \greedyLEGO{} limits the set of available tasks at a given assignment iteration to the tasks that belong to active build steps.

\section{Plan Execution and Collision Avoidance}\label{sec:rvo_route_planning}

% New setting is continuous space + time
% We could plan trajectories ahead of time, but this might break down under uncertainty, delays, etc.
% Rather, we adopt an online navigation strategy base on highways, visibility graphs, and reactive collision avoidance.

With a staging plan defined, all transport tasks allocated to robots and robot teams, and the transport units configured, we now need to execute the construction plan. This requires robots to move through the environment, collecting, transporting, and depositing their cargo, all while avoiding collision with each other and the various assemblies under construction throughout the factory.

At any given moment in the construction process, a subset of the assembly build phases are active. Each active staging area is treated as a ``soft'' obstacle for all agents that are not directly involved in the activities of that staging area. More precisely, an agent should only enter an active staging area if (a) the agent's current task requires it to enter the staging area, and (b) the build step associated with the task is active. Otherwise, an agent may only enter a staging area if ``pushed'' into the staging area by another agent. Recall that the staging area for each build phase is defined as the minimum radius cylinder that fully encloses the assembly's current bounding cylinder, all dropoff zones associated with the build step, and the staging area of the previous build phase. Thus, when a non-terminal build phase is completed, a larger staging area becomes active. The robot fleet must therefore navigate through an environment where virtual obstacles appear and disappear over time.

One approach would be to precompute an execution plan, consisting of the trajectories of all agents from time zero to the completion of the project. Such an approach would be analogous to the pre-execution route planning method used in \cite{Brown2020a}. A pre-computation approach is attractive because, with an appropriate global optimization method, it allows the possibility of finding a makespan-optimal execution plan (that is, optimal with respect to the construction plan). \citet{Li2020a} propose a prioritized multi-robot trajectory optimization scheme that could be applicable here. However, a long-horizon plan can easily break down due to delays caused by unforeseen disturbances in the environment. Moreover, a precomputation approach would need to account for the appearances and disappearances of virtual obstacles, which in turn depend on the times at which different tasks are completed.

Instead of precomputing an execution plan, we propose a distributed online navigation strategy wherein each agent follows a reactive velocity control policy where we use the term \emph{agent} to mean either a robot or a transport unit.
% Each agent is modeled as a cylinder (Robots are cylindrical in the first place. Each transport unit is simply overapproximated with its minimum radius bounding cylinder). 
The reactive policy consists of three layers. The first layer is a simple switching controller that plans a nominal velocity vector meant to move the agent toward its goal while avoiding active staging areas that the agent should not enter.
The second layer of the reactive policy adds a dispersion component to the nominal velocity based on weighted, pairwise repulsive artificial potential fields.
The third and final layer is a collision-avoidance controller that computes an updated velocity vector if the preferred velocity vector would lead to collision with other agents. The output of this final layer is the commanded velocity.

\begin{algorithm}[htb!]
  \small
  \caption{The three-level distributed velocity controller.}\label{alg:velocity_controller}
  \begin{algorithmic}[1]
    \State \textbf{Input:}
        \State \quad $\x$: Current positions and goals all agents
    \State \textbf{Output:}
        \State \quad $\vcom$: Collision-free velocity command for each agent
    \vspace{0.5em}
    \Procedure{VelocityController}{$\x$}
      \State $\vnom \gets \textsc{\TangentBugPolicy}(\x)$
      \State $\vpref \gets \textsc{\DispersionProtocol}(\x,\vnom)$
      \State $\vcom \gets \textsc{\RVO}(\x,\vpref)$
    \EndProcedure
  \end{algorithmic}
\end{algorithm}

\subsection{Level 1: Modified Tangent Bug Algorithm}\label{sec:tangent_bug}

\begin{algorithm*}[htb!]
  \small
  \caption{The modified tangent bug controller.}\label{alg:tangent_bug}
  \begin{algorithmic}[1]
    \State \textbf{Input:}
        \State \quad $\bugpos$: Current position of the agent
        \State \quad $\buggoal$: Goal location of the agent
    \State \textbf{Output:}
        \State \quad $\waypoint$: Intermediate waypoint computed to avoid staging area obstacles
        \State \quad $mode$: Navigation mode 
    \vspace{0.5em}
    \Procedure{ModifiedTangentBug}{$\bugpos$,$\buggoal$}
      \State $\target \gets$ first obstacle intersected by ray $\bugline$ 
      \If{$\target \neq nothing$}
        \State $\waypoint \gets$ point at which $\bugline$ first intersects $\target$
        \State $d \gets$ signed distance from $\bugpos$ to boundary of $\target$
        \If{$d \approx 0$} \Comment $\bugpos$ is on boundary of $\target$
          \State $mode \gets \textsc{move\_ccw\_along\_boundary}$
        \ElsIf{$d > 0$} \Comment $\bugpos$ is outside of $\target$
          \If{$d > \bugradius$}
            \State $mode \gets \textsc{move\_toward\_waypoint}$
          \Else
            \State $o \gets$ first obstacle intersected by ray $\bugpos \rightarrow \waypoint$ 
            % \State $\waypoint \gets$ right hand tangent point on $\target$ boundary from $\bugpos$
            \If{$o = nothing$}
              \State $mode \gets \textsc{move\_toward\_right\_hand\_tangent\_point}$
            \Else
              \State $mode \gets \textsc{move\_toward\_waypoint}$
            \EndIf
          \EndIf
        \ElsIf{$d < 0$} \Comment $\bugpos$ is inside of $\target$
          \State $mode \gets \textsc{exit\_target}$
        \EndIf
      \Else
        \State $\waypoint \gets \buggoal$
        \State $mode \gets \textsc{move\_toward\_waypoint}$
      \EndIf
    \EndProcedure
  \end{algorithmic}
\end{algorithm*}

Given the current set of staging area obstacles, each robot computes its own nominal velocity using a variation of the Tangent Bug algorithm \citep{Kamon1998}. The agent's waypoint is initialized as its goal location. If the straight path from the agent's current position toward the waypoint is unobstructed up to some lookahead distance, the nominal velocity is simply set to a vector pointing along that path. If the robot is far from the goal, the vector's magnitude is the maximum permissible speed of that agent, as defined by \cref{eqn:transport_unit_speed_limit}. If the robot is within a single time step of reaching the goal, the velocity is scaled so that the robot will not overshoot the goal.

If the path from start to waypoint is blocked by one or more obstacles at a closer proximity than the lookahead distance, the closest of these obstacles is designated as the target. The waypoint is set to the right-hand tangent point (i.e., the robot aims to ``skim'' the obstacle by passing along its right side) of the circle created by inflating the target by the agent's own radius. If the path to this new waypoint is obstructed by another obstacle, the waypoint is instead set to the first point on the inflated target's boundary that the agent would reach if it were to travel straight toward its goal location. The nominal velocity is then set as a vector of maximum permissible magnitude in the direction of the waypoint.

If the agent's position is within some $\epsilon$ of the inflated target's boundary, the agent selects a nominal velocity that will move it along the boundary in the counter-clockwise direction. When the agent reaches a point on the boundary at which the target no longer obstructs a straight path to the agent's goal, the target is discarded and the agent selects a new waypoint.

As agents switch tasks and new build steps become active, an agent will occasionally find itself within a staging area in which it is not permitted to be. In this case, the agent simply selects a velocity that follows the shortest path to the outside of the staging area. In the event that the agent is at the exact center of the staging area, its exit path points in the direction of the agent's goal.

Our modified tangent bug algorithm always leads to counter-clockwise detours around obstacles. This helps to reduce congestion that would occur if two agents tried to navigate around the same obstacle in opposite directions. That said, the nominal velocities computed by the tangent bug policy might lead to collisions if executed directly by the agents. %Hence, the tangent bug layer of each agent's control policy is followed by a ``collision-avoidance'' layer.

\subsection{Level 2: Prioritized Dispersion Protocol}\label{sec:dispersion_protocol}
The second level of the velocity controller defines active agents as follows: a transport unit is designated as active if it is carrying cargo that belongs to an active build step. A robot is designated as active if the next task in the robot's itinerary is to join a transport unit whose cargo (a) is available for pickup and (b) belongs to an active build step.

When an active agent reaches the staging circle within which its goal lies, the agent may enter immediately. Inactive agents, on the other hand, must wait outside of the circle. When multiple inactive agents are waiting outside of a circle, there may not be enough room for an active agent to make its way through the crowd. Intuitively, inactive agents need to make room for an active agent when the active agent needs to pass.

The prioritized dispersion protocol causes inactive agents to move away from other agents when an active agent is close. This allows active agents to ``push through'' crowds of inactive agents. The dispersion protocol is based on virtual pairwise repulsive potential fields. Each inactive agent is subject to repulsive fields emanating from other nearby agents. A dynamic priority scheme (\cref{alg:rvo_priority_scheme}) is used to determine higher priority (lower $\alpha$ is higher priority). An agent with higher priority is not affected by the repulsive fields of other agents.
The repulsive force exerted by agent $j$ on agent $i$ is defined by 
\begin{align}
    \potentialfield_1(\x_i, \x_j,\agentRadius_i,\agentRadius_j,\bufferRadius_j) &= \max(0, \bufferRadius_j + \agentRadius_i + \agentRadius_j \label{eqn:cone_potential} \\
    &\hspace{8em} - \Vert \x_i - \x_j \Vert), \nonumber \\
    \potentialfield_2(\x_i, \x_j,\agentRadius_i,\agentRadius_j,\bufferRadius_j) &= \max(0, 1 / (\Vert \x_i - \x_j \Vert - \bufferRadius_j) \label{eqn:barrier_potential} \\
    &\hspace{8em} - 1 / (\agentRadius_i + \agentRadius_j), \nonumber \\
    \potentialfield(\cdot) &= \potentialfield_1(\cdot) + \potentialfield_2(\cdot), \label{eqn:composite_potential} \\
    f &= \nabla_{\x_i} \potentialfield(\x_i, \x_j,\agentRadius_i,\agentRadius_j,\bufferRadius_j) \label{eqn:repulsive_force}
\end{align}
where $\x_i$ and $\x_j$ denote the agents' position vectors, $\agentRadius_i$ and $\agentRadius_j$ denote the radii of the agents' bounding spheres, $\bufferRadius_j$ denotes the field radius of agent $j$, \cref{eqn:cone_potential} encodes a cone-shaped potential $\potentialfield_1$, \cref{eqn:barrier_potential} encodes a log barrier-shaped potential $\potentialfield_2$, \cref{eqn:composite_potential} defines the overall potential $\potentialfield$ as the sum of the cone and barrier potentials, and \cref{eqn:repulsive_force} defines the repulsive force $f$ as the gradient of the potential field with respect to $\x_i$.

The field radius $\bufferRadius_j$ determines how far the potential field extends from agent $j$. For a small value of $\bufferRadius_j$, agent $j$ only exerts a repulsive force on agents that are very close to it. Increasing $\bufferRadius_j$ has the effect of expanding the neighborhood in which other agents are affected by the repulsive force from $j$. The value of $\bufferRadius_j$ depends inversely on the distance from agent $j$ to the nearest active agent, according to
\begin{align}
  & d_j = \min_{k \in active\_agents} \Vert \x_j - \x_k \Vert - (\agentRadius_k + \agentRadius_j), \label{eqn:dist_to_nearest_active_agent} \\
  & \bufferRadius_j = \min(\maxbufferRadius,\bufferMultiplier / d_j), \label{eqn:define_buffer_radius}
\end{align}
where $d_j$ denotes the distance from agent $j$ to the nearest active agent, $\bufferRadius_{\textsc{max}}$ is an upper bound on the field radius, and $\bufferMultiplier$ is a scaling hyperparameter (we use $\maxbufferRadius=2.5\robotradius$ and $\bufferMultiplier=\robotradius$). If agent $j$ is an active agent, its field radius is equal to $\maxbufferRadius$.

Note that the force exerted on $i$ by $j$ is not necessarily equal in magnitude to the force exerted on $j$ by $i$. As previously noted, active agents are not affected by the potential fields. Between inactive agents, the field radius will be larger for the agent that is closer to an active agent. 

The overall virtual force experienced by agent $i$ is the sum of the forces exerted by all other agents within its vicinity. The preferred velocity of agent $i$ is computed by blending the nominal velocity with the virtual force, and clipping the resulting velocity vector if its magnitude exceeds the maximum permissible speed:
\begin{align}
  & \hat{\vel} = a \vnom - b  \sum_j \nabla_{\x_i}\potentialfield(\x_i, \x_j,\agentRadius_i,\agentRadius_j,\bufferRadius_j) \\
  & \vpref = \frac{\vel}{\Vert \hat{\vel} \Vert} \min(\vel_{\textsc{max}}, \Vert \hat{\vel} \Vert)
\end{align}%
where $a$ and $b$ are blending coefficients. We use $a=1$ and $b=1$ in our experiments.

\subsection{Level 3: Generalized RVO with Dynamic Prioritization}\label{sec:prioritized_RVO}
The final layer of the velocity controller is based on reciprocal velocity obstacles (RVO).
A velocity obstacle is created by translating the relative position vector between two robots and the desired velocity vector of the first robot into a set of two inequality constraints on the velocity of second robot. Any velocity vector in the second robot's velocity envelope that satisfies either of these constraints will not lead to collision with the first robot. Reciprocal velocity obstacles extend the velocity obstacle concept by having pairs of robots share responsibility for avoiding collision with each other. Generalized reciprocal velocity obstacles extend this notion further by allowing two agents to share collision-avoidance responsibility unevenly. The parameter $\alpha^{i}_{j} \in [0,1]$ denotes the share of the responsibility that agent $i$ takes to avoid collision with agent $j$ (agent $j$'s share of the responsibility is $\alpha^j_i = 1 - \alpha^i_j$) \citep{VanDenBerg2008}.

We use generalized RVO with a dynamic priority scheme (\cref{alg:rvo_priority_scheme}) that assigns to each agent its own non-negative $\alpha$-value. Each time any agent completes a task, the $\alpha$-values of all agents are recomputed. For any two agents $i$ and $j$, we compute $\alpha^i_j = \alpha_i / (\alpha_i + \alpha_j)$. If $\alpha_i = \alpha_j = 0$, we simply set $\alpha^i_j$ to $0.5$. The priority scheme is designed to prioritize robots and transport units that are engaged in active build phases, so that they can more easily push past other agents who are waiting for their own build phases to begin. Within the active build phases, transport units are given higher priority than unladen robots because they are ``ahead'' of the robots in completing their tasks (the unladen robots are on their way to form transport units). 

\begin{algorithm}[htb!]
\small
  \caption{The dynamic priority scheme for setting an agent's $\alpha$-value (priority).}\label{alg:rvo_priority_scheme}
  \begin{algorithmic}[1]
    \State \textbf{Input:}
        \State \quad $\rvoagent$: Agent (robot or transport unit) whose priority is to be computed
    \State \textbf{Output:}
        \State \quad $\alpha$: Priority value determining agent's collision-avoidance responsibility
    \vspace{0.5em}
    \Procedure{SetAlphaValue}{$\rvoagent$}
      \If{$\rvoagent$ is a transport unit}
        \If{task is $\FormTransportUnit$ or $\DepositCargo$}
          \State $\alpha \gets 0$ \Comment must remain stationary
        \Else \Comment task is $\TransportUnitGo$
          \State $cargoScale \gets \textsc{cargoID} / (10 \cdot \textsc{maxCargoID})$ %\Comment{provide priority based on cargo}
          \If{task's build phase is active}
            \State $\alpha \gets 0 + cargoScale$
          \Else
            \State $\alpha \gets 1$
          \EndIf
        \EndIf
      \Else \Comment $\rvoagent$ is an unladen robot
        \If{task's build phase is active}
          \If{task's cargo is ready for pickup}
            \State $\alpha \gets 0.1$
          \Else
            \State $\alpha \gets 0.5$
          \EndIf
        \Else
          \State $\alpha \gets 1$
        \EndIf
      \EndIf
    \EndProcedure
  \end{algorithmic}
\end{algorithm}

\subsection{Task Swapping}\label{sec:avoiding_deadlock}

In some cases, a member of a transport unit is unable to reach its carrying position because other members of the robot team are already waiting in their assigned pickup locations. When such deadlocks occur, we simply allow the stuck robot to swap tasks with the nearest team member that is closer to the stuck robot's goal than the stuck robot.

\subsection{Sit-And-Wait Subroutine for Inactive Agents}\label{sec:sit_and_wait}

In experiments with the distributed execution controller described above, we find that inactive robots tended to oscillate, pushing each other back and forth while tightly gathered around a staging area. To avoid this needless dancing, we added a sit-and-wait subroutine that sets the nominal velocity to zero for inactive agents within a specific distance from their destinations. With this feature enabled, inactive agents within stopping range of their goal will not move unless the $\DispersionProtocol$ or $\RVO$ policy layers require it to deviate from its zeroed nominal velocity. This prevents most of the undesirable oscillations observed without the sit-and-wait feature, though some oscillation is still observed when an inactive robot far from its goal tries to move through a large group of other waiting inactive agents.

\section{Demonstrations}\label{sec:LEGO_demos}

We demonstrate our system's performance in a simulated environment, which is shown in \cref{fig:tractor_demo_sequence}. The initial robot positions are drawn from a uniform distribution over a grid of locations around the center of the environment. The final assembly staging area is always at the origin. All assemblies are constructed in the air above the robots. When a transport unit deposits its cargo, the cargo rises into the air until it reaches its pre-lift-into-place configuration. Once the cargo reaches this location, the \DepositCargo task is complete and then the item is moved into its location within the assembly. The initial locations of all raw materials are placed at random locations outside of the staging areas dictated by the staging plan. 

All simulator code is written in Julia \citep{Bezanson2017} and the optimization was performed with Gurobi \citep{gurobi}. Rendering is done by MeshCat.\footnote{\url{https://github.com/rdeits/MeshCat.jl}} We use Julia's PyCall package to access the Python bindings to a modified version of the RV02 Library\footnote{The RVO2 C++ Library is available at \url{https://gamma.cs.unc.edu/RVO2}} (our modified RVO2 library incorporates the $\alpha$ prioritization levels described in \cref{sec:prioritized_RVO}).  

\begin{figure*}[tb]
    \centering
    \resizebox{\linewidth}{!}{%
        \input{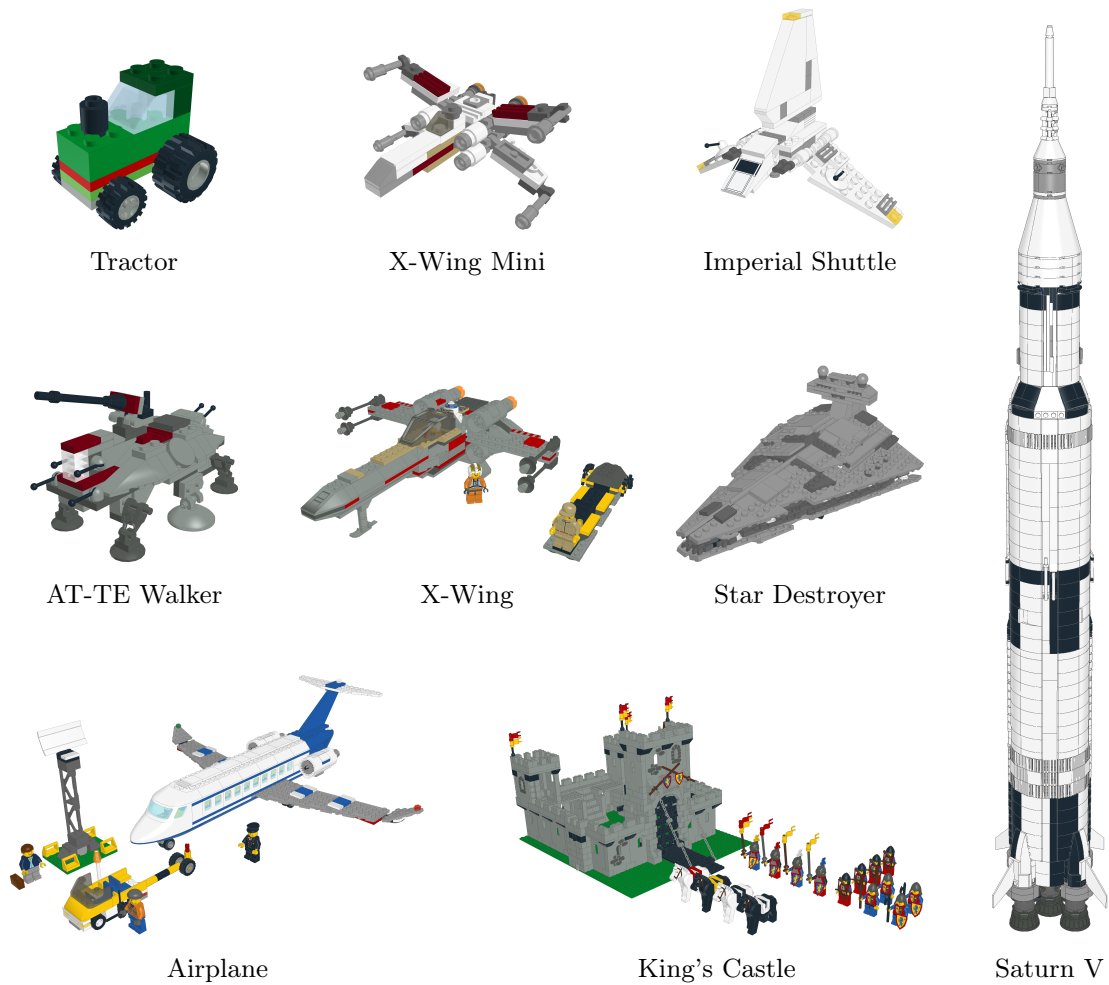}
    }
    \caption{Nine assemblies we use to test our algorithm with different levels of complexity.}
    \label{fig:five_assemblies}
\end{figure*}

We demonstrate our framework on nine different assemblies which span different complexities in terms of number of parts and assemblies (\cref{fig:five_assemblies}):
\begin{itemize}
  \item Tractor: The tractor project that has been used as a running example throughout this paper. This model is based on \LEGO{} model 10708, \emph{Green Creativity Box}. It consists of $20$ pieces organized into $8$ assemblies.
  \item X-Wing Mini: Based on \LEGO{} model 30051, \emph{X-wing Fighter - Mini}, from the \LEGO{} Star Wars collection. It consists of 61 parts organized into $12$ assemblies.
  \item Imperial Shuttle: Based on \LEGO{} model 4494, \emph{Imperial Shuttle - Mini}, from the \LEGO{} Star Wars collection. It consists of $84$ parts organized into $5$ assemblies.
  \item AT-TE Walker: Based on \LEGO{} model 20009, \emph{AT-TE Walker - Mini}, from the \LEGO{} Star Wars collection. AT-TE Walker consists of $100$ parts organized into $22$ assemblies.
  \item X-Wing: Based on \LEGO{} model 7140, \emph{X-wing Fighter} from the \LEGO{} Star Wars collection. X-Wing is a more complex assembly than X-Wing Mini, with 309 parts organized into 28 assemblies.
  \item Airplane: Based on \LEGO{} model 3181, \emph{Passenger Plane}, from the \LEGO{} City collection. It consists of $326$ parts organized into $28$ assemblies.
  \item Star Destroyer: Based on \LEGO{} model 8099, \emph{Midi-Scale Imperial Star Destroyer} from the \LEGO{} Star Wars collection. It consists of $418$ parts organized into 11 assemblies.
  \item King's Castle: Based on \LEGO{} model 6080, \emph{King's Castle}. This model consists of $761$ parts organized into $70$ assemblies.
  \item Saturn V: Based on \LEGO{} model 21309, \emph{NASA Apollo Saturn V}. The Saturn V rocket has $1845$ pieces organized into $306$ assemblies. 
\end{itemize}

\subsection{Full Stack Simulations}\label{sec:full_stack_LEGO_results}
\begin{figure*}[tb]
  \centering
  \begin{subfigure}{0.45\textwidth}
    \centering
    \includegraphics[width=\textwidth]{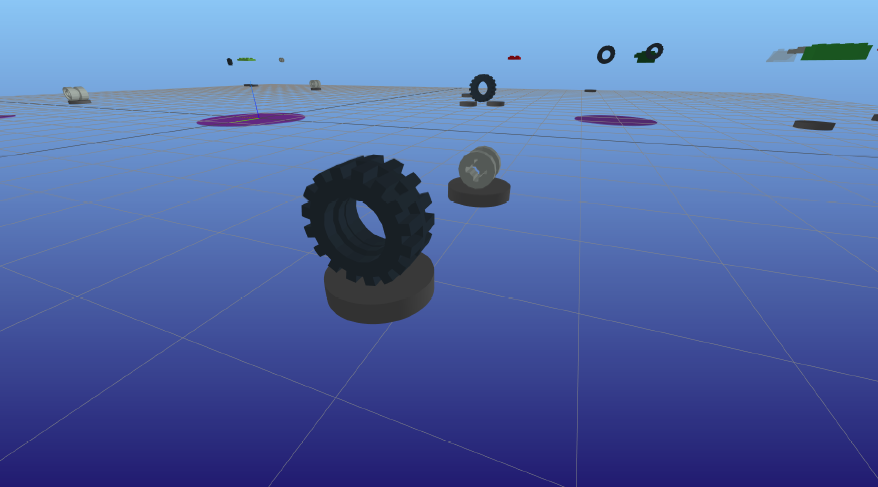}
    \caption{}
    \label{fig:tractor_demo_a}
  \end{subfigure}%
  \hfill
  \begin{subfigure}{0.45\textwidth}
    \centering
    \includegraphics[width=\textwidth]{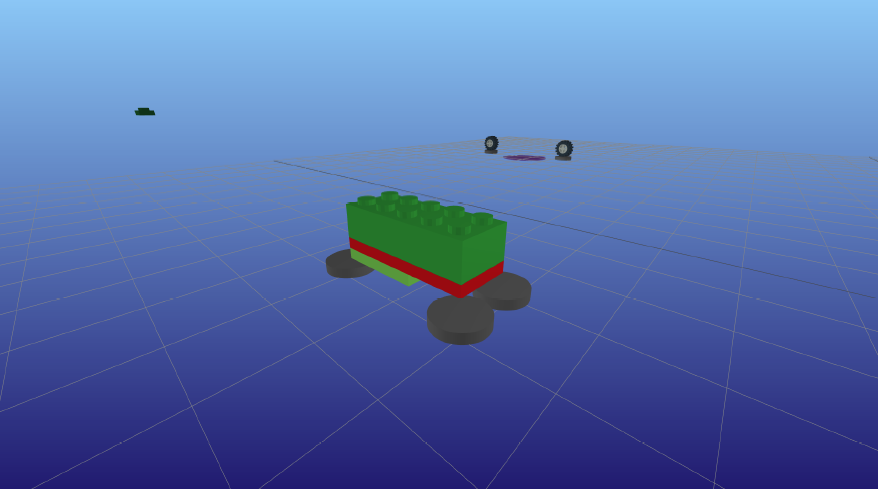}
    \caption{}
    \label{fig:tractor_demo_b}
  \end{subfigure}

  \vspace{1em}

  \begin{subfigure}{0.45\textwidth}
    \centering
    \includegraphics[width=\textwidth]{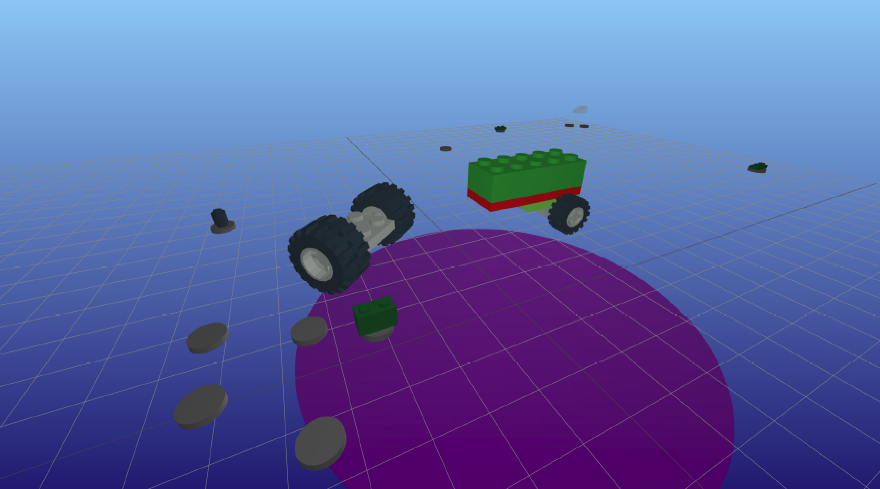}
    \caption{}
    \label{fig:tractor_demo_c}
  \end{subfigure}%
  \hfill
  \begin{subfigure}{0.45\textwidth}
    \centering
    \includegraphics[width=\textwidth]{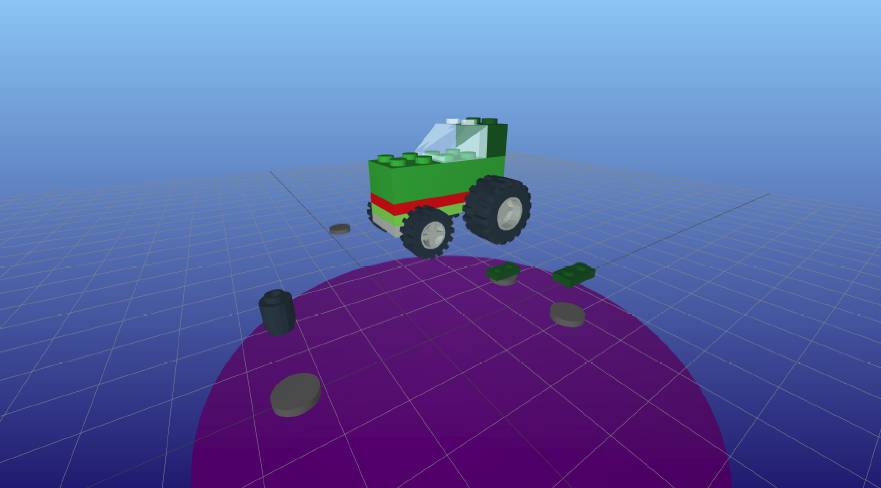}
    \caption{}
    \label{fig:tractor_demo_d}
  \end{subfigure}

  \caption{Screenshots from the Tractor construction in the simulated environment: (a) Components of a tire assembly are carried by two robots; (b) The chassis is transported by a team of four robots; (c) The rear axle is lifted into place, while some of the final components are seen on board robots in the background; (d) The final pieces of the Tractor assembly are lifted into place.}
  \label{fig:tractor_demo_sequence}
\end{figure*}

\begin{figure*}[tb]
  \centering
  \begin{subfigure}{0.45\textwidth}
    \centering
    \includegraphics[width=\textwidth]{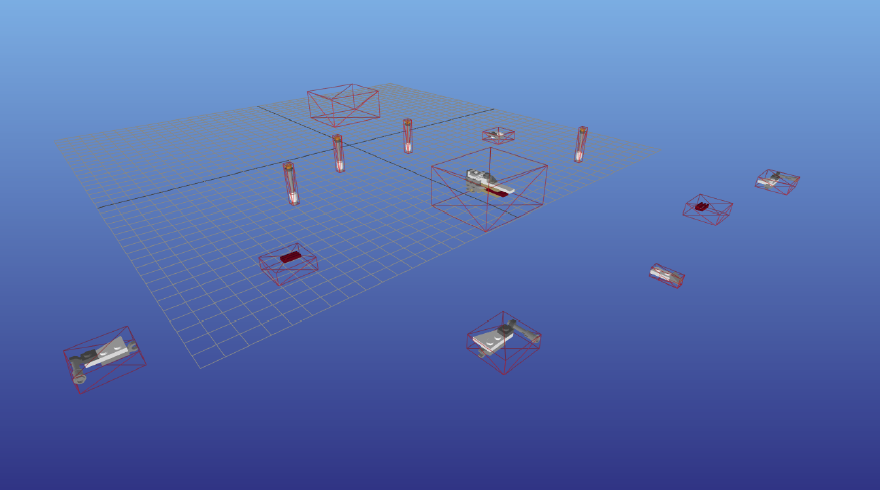}
    \caption{}
    \label{fig:xwing_demo_a}
  \end{subfigure}%
  \hfill
  \begin{subfigure}{0.45\textwidth}
    \centering
    \includegraphics[width=\textwidth]{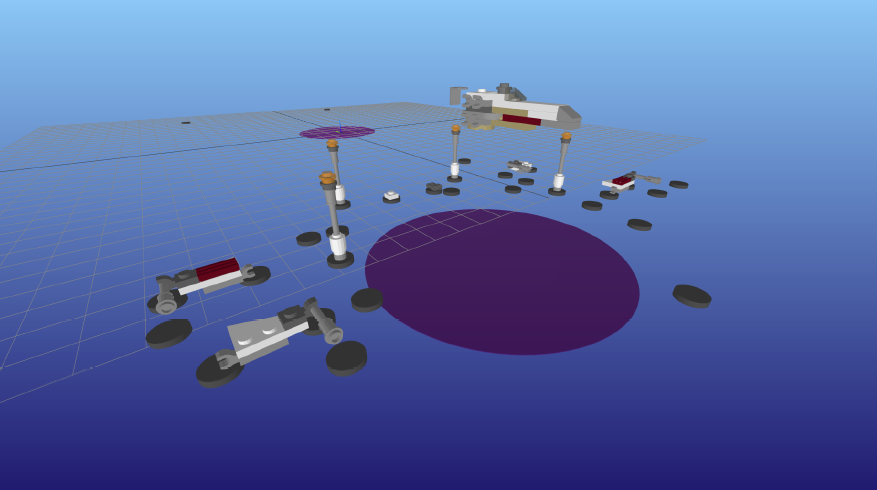}
    \caption{}
    \label{fig:xwing_demo_b}
  \end{subfigure}

  \vspace{1em}

  \begin{subfigure}{0.45\textwidth}
    \centering
    \includegraphics[width=\textwidth]{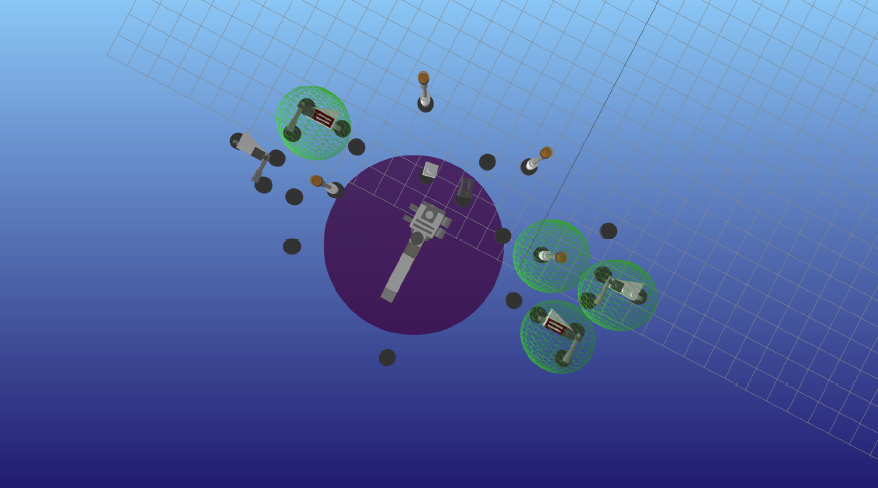}
    \caption{}
    \label{fig:xwing_demo_c}
  \end{subfigure}%
  \hfill
  \begin{subfigure}{0.45\textwidth}
    \centering
    \includegraphics[width=\textwidth]{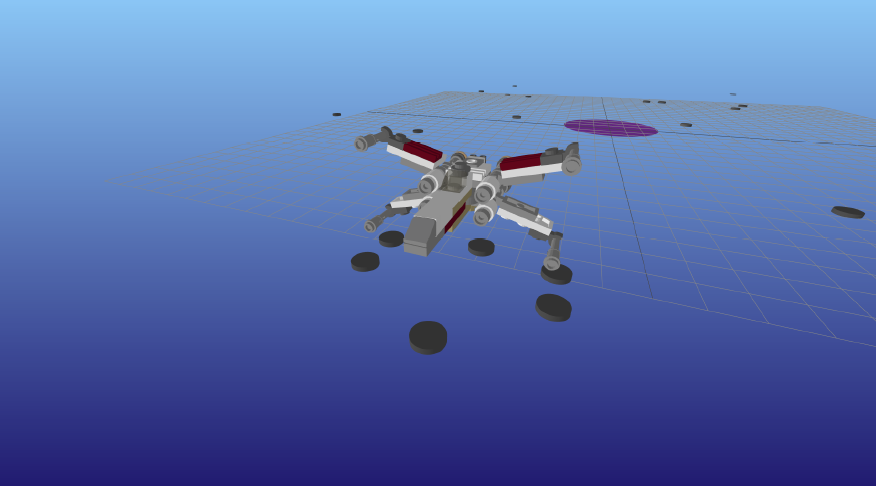}
    \caption{}
    \label{fig:xwing_demo_d}
  \end{subfigure}

  \caption{Screenshots from the X-Wing Mini construction in the simulated environment: (a) The bounding hyperrectangles of the subassemblies are shown at the assembly build locations (this snapshot was taken prior to the beginning of the simulation). (b) Essentially all of the robots are congregated around the penultimate staging area, waiting for their turn to enter; (c) A bird's eye view shows the crowding from a different angle. The bounding hyperspheres for some of the transport units are shown to convey the tightness of the crowding; (d) A group of six robots waits in a hexagonal carrying formation as the completed assembly is lowered into its carrying configuration.}
  \label{fig:xwing_mini_demo_sequence}
\end{figure*}

For each assembly, we ran a full stack simulation with different parameters. The execution for all parts of the simulation occurred on a system using an Intel$^{\text{\textregistered}}$ Core$\text{\texttrademark}$ i9-9900KF processor and $\SI{64}{\giga\byte}$ of RAM. 
Images from the Tractor and X-Wing Mini projects are shown in \cref{fig:tractor_demo_sequence} and \cref{fig:xwing_mini_demo_sequence} respectively. Additional simulations and interactive visualizations of the assembly processes are available in our repository.\footnote{\url{https://github.com/sisl/ConstructionBots.jl}}

\Cref{tab:demo_results} reports metrics from the simulation runs on all nine assemblies and various numbers of robots. The model names are listed with the number of parts and assemblies. The simulations in \cref{tab:demo_results} use \greedyLEGO{} for task assignments and perform execution using the three components of the collision avoidance strategy discussed, \TangentBugPolicy{}, \DispersionProtocol{}, and \RVO{}. The following metrics are used in \cref{tab:demo_results} and throughout this section:
\begin{itemize}
  \item Transport Unit Configuration Time (T.U. Config): The total time spent configuring all transport units (\cref{sec:configure_transport_units}).
  \item Staging Plan Generation Time (Staging Plan): The time spent generating the global staging plan (\cref{sec:construct_staging_plan}). 
  \item Task Allocation Time (Assignment):  The time spent forming coalitions and allocating tasks (\cref{sec:collaborative_task_allocation}).
  \item Predicted Makespan: This value is the predicted simulation time required to complete the project. The predicted makespan is a result of the output of the task assignment solution and this makespan is the metric used for the optimization and does not involve collision avoidance maneuvers.
  \item Execution Makespan: The makespan of the actual execution run (i.e., the simulation time required to complete the project). This value is longer than the predicted makespan due to the avoidance of staging circles and the collision avoidance logic.
  \item Execution Runtime (Runtime): The amount of wallclock time required to run the simulator to project completion. This time does not include the preprocessing steps.  
\end{itemize}

The total preprocessing time of the three components listed using \greedyLEGO{} assignment for each of the projects is less than three minutes. The entire construction plan for the two smallest projects is computed in less than a second, whereas the largest project (Saturn V with $250$ robots) takes approximately $2$ minutes and $49$ seconds. Task allocation is the largest preprocessing computational burden. Configuring transport units and the staging plan generation are independent of the number of robots and both are fast, taking less than three seconds for the largest project.

The distributed execution strategy successfully completes all projects. However, there are no guarantees to prevent deadlock. The combination of the control strategies we discussed can still cause deadlocks, especially as the buffer size between staging areas is reduced. We discuss potential improvement ideas in \cref{sec:LEGO_discussion}. We also note that it is clear from \cref{tab:demo_results} that the simulator runtime scales quite poorly with project size. This runtime is an artifact of our implementation strategy and not a factor of our distributed execution strategy. We currently are processing all agent and component actions and updates on a single thread. The runtime would improve by taking advantage of a multi-threaded simulation environment and implementing more efficient data structures.

\begin{table*}
% \tiny 
% \scriptsize
% \footnotesize 
% \small
\normalsize
\centering
\ra{1.11}
\setlength{\tabcolsep}{3.5pt}
\caption{Results for full planner and simulation stack. Task allocation is performed by \greedyLEGO{}, and execution is performed using \TangentBugPolicy{}+\DispersionProtocol{}+\RVO{}}
\begin{adjustbox}{max width=\textwidth}
\begin{tabular}{@{}llccccccc@{}}
    \toprule
    \multicolumn{2}{l}{Model (Parts$/$Assemblies)} &  \multicolumn{3}{c}{Preprocessing (\SI{}{\second})} && \multicolumn{2}{c}{Makespan (\SI{}{\second})} & \multirow{2}{*}{Runtime (\SI{}{\second})} \\
    \cmidrule{3-5} \cmidrule{7-8}
    & \quad$\#$ Robots & T.U. Config & Staging Plan & Assignment && Predicted & Execution \\
    \midrule
    \multicolumn{2}{l}{Tractor $(20 / 8)$} \\
    & \quad$\phantom{0}\phantom{0}5$    & $0.03$ & $0.05$ & $\phantom{0}\phantom{0}0.1$ && $\phantom{0}27.4$ & $\phantom{0}35.0$ & $\phantom{0}\phantom{0}\phantom{0}1.4$ \\
    & \quad$\phantom{0}10$              & $0.03$ & $0.05$ & $\phantom{0}\phantom{0}0.1$ && $\phantom{0}16.1$ & $\phantom{0}18.8$ & $\phantom{0}\phantom{0}\phantom{0}1.7$ \\
    & \quad$\phantom{0}15$              & $0.03$ & $0.05$ & $\phantom{0}\phantom{0}0.1$ && $\phantom{0}11.9$ & $\phantom{0}19.0$ & $\phantom{0}\phantom{0}\phantom{0}2.8$ \\
    \multicolumn{2}{l}{X-Wing Mini $(61 / 12)$} \\
    & \quad$\phantom{0}15$              & $0.08$ & $0.1$ & $\phantom{0}\phantom{0}0.4$ && $\phantom{0}31.2$ & $\phantom{0}43.9$ & $\phantom{0}\phantom{0}\phantom{0}8.8$ \\
    & \quad$\phantom{0}20$              & $0.08$ & $0.1$ & $\phantom{0}\phantom{0}0.4$ && $\phantom{0}23.3$ & $\phantom{0}33.9$ & $\phantom{0}\phantom{0}\phantom{0}9.3$ \\
    & \quad$\phantom{0}25$              & $0.08$ & $0.1$ & $\phantom{0}\phantom{0}0.4$ && $\phantom{0}20.1$ & $\phantom{0}34.1$ & \phantom{0}\phantom{0}12.8 \\
    \multicolumn{2}{l}{Imperial Shuttle $(84 / 5)$} \\
    & \quad$\phantom{0}15$              & $0.1$ & $0.1$ & $\phantom{0}\phantom{0}0.4$ && $\phantom{0}55.9$ & $\phantom{0}66.0$ & $\phantom{0}\phantom{0}12.6$ \\
    & \quad$\phantom{0}20$              & $0.1$ & $0.1$ & $\phantom{0}\phantom{0}0.4$ && $\phantom{0}43.8$ & $\phantom{0}57.2$ & $\phantom{0}\phantom{0}15.2$ \\
    & \quad$\phantom{0}25$              & $0.1$ & $0.1$ & $\phantom{0}\phantom{0}0.4$ && $\phantom{0}34.0$ & $\phantom{0}54.3$ & $\phantom{0}\phantom{0}18.4$ \\
    \multicolumn{2}{l}{AT-TE Walker $(100 / 22)$} \\
    & \quad$\phantom{0}25$              & $0.2$ & $0.2$ & $\phantom{0}\phantom{0}0.7$ && $\phantom{0}37.1$ & $\phantom{0}48.2$ & $\phantom{0}\phantom{0}19.0$ \\
    & \quad$\phantom{0}35$              & $0.2$ & $0.2$ & $\phantom{0}\phantom{0}0.7$ && $\phantom{0}29.9$ & $\phantom{0}38.9$ & $\phantom{0}\phantom{0}22.8$ \\
    & \quad$\phantom{0}45$              & $0.2$ & $0.2$ & $\phantom{0}\phantom{0}0.7$ && $\phantom{0}23.4$ & $\phantom{0}31.9$ & $\phantom{0}\phantom{0}26.2$ \\
    \multicolumn{2}{l}{X-Wing $(309 / 28)$} \\
    & \quad$\phantom{0}40$              & $0.7$ & $0.4$ & $\phantom{0}\phantom{0}3.4$ && $144.5$ & $179.7$ & $\phantom{0}160.7$ \\
    & \quad$\phantom{0}50$              & $0.7$ & $0.4$ & $\phantom{0}\phantom{0}3.4$ && $124.8$ & $155.4$ & $\phantom{0}178.1$ \\
    & \quad$\phantom{0}60$              & $0.7$ & $0.4$ & $\phantom{0}\phantom{0}3.5$ && $106.2$ & $140.2$ & $\phantom{0}216.3$ \\
    \multicolumn{2}{l}{Airplane $(326 / 28)$} \\
    & \quad$\phantom{0}40$              & $0.7$ & $0.3$ & $\phantom{0}\phantom{0}3.7$ && $207.6$ & $268.1$ & $\phantom{0}226.8$ \\
    & \quad$\phantom{0}50$              & $0.7$ & $0.3$ & $\phantom{0}\phantom{0}3.7$ && $166.8$ & $220.5$ & $\phantom{0}245.5$ \\
    & \quad$\phantom{0}60$              & $0.7$ & $0.3$ & $\phantom{0}\phantom{0}3.8$ && $144.5$ & $184.7$ & $\phantom{0}261.1$ \\
    \multicolumn{2}{l}{Star Destroyer $(418 / 11)$} \\
    & \quad$\phantom{0}65$              & $0.6$ & $0.3$ & $\phantom{0}\phantom{0}4.2$ && $113.7$ & $149.9$ & $\phantom{0}233.0$ \\
    & \quad$\phantom{0}75$              & $0.6$ & $0.3$ & $\phantom{0}\phantom{0}4.3$ && $104.9$ & $136.3$ & $\phantom{0}254.4$ \\
    & \quad$\phantom{0}85$              & $0.6$ & $0.3$ & $\phantom{0}\phantom{0}4.3$ && $\phantom{0}90.0$ & $117.5$ & $\phantom{0}259.7$ \\
    \multicolumn{2}{l}{King's Castle $(761 / 70)$} \\
    & \quad$\phantom{0}75$              & $1.2$ & $0.6$ & $\phantom{0}15.4$ && $291.8$ & $317.0$ & $\phantom{0}719.0$ \\
    & \quad$125$                        & $1.2$ & $0.6$ & $\phantom{0}16.7$ && $196.8$ & $224.0$ & $\phantom{0}986.9$ \\
    & \quad$175$                        & $1.2$ & $0.6$ & $\phantom{0}19.1$ && $148.8$ & $186.9$ & $1492.9$ \\
    \multicolumn{2}{l}{Saturn V $(1845 / 306)$} \\
    & \quad$150$                        & $2.9$ & $3.0$ & $148.8$ && $334.6$ & $391.8$ & $4457.0$ \\
    & \quad$200$                        & $2.9$ & $3.0$ & $156.2$ && $281.7$ & $324.3$ & $5457.8$ \\
    & \quad$250$                        & $2.9$ & $3.0$ & $163.2$ && $257.9$ & $298.7$ & $6849.4$ \\
    \bottomrule
\end{tabular}
\label{tab:demo_results}
\end{adjustbox}
\end{table*}

\subsection{Task Allocation Comparison}\label{sec:LEGO_task_assignment_results}
The \greedyLEGO{} task allocation produced feasible solutions. To further evaluate the quality of the solutions, we compared \greedyLEGO{} to the \SparseAdjacencyMILP{} and \SparseAdjacencyMILP{} with a \greedyLEGO{} warm-start task allocation methods. An optimizer time limit was used with \SparseAdjacencyMILP{}. All but Tractor with $15$ robots reached the optimizer time limit before finding an optimal solution. We used a time limit of \SI{6000}{\second} for all problems except King's Castle and Saturn V where we used \SI{12000}{\second}. On Saturn V, the \textsc{Spar\-seAdjacencyMILP} formulation with a warm-start was unable to find a feasible solution to improve upon the \greedyLEGO{} solution with a \SI{12000}{\second} optimizer time limit. Therefore, we did not include Saturn V in the table. 
\SparseAdjacencyMILP{} with and without a warm-start were run with the Gurobi \textsc{MIPFocus} parameter set to focus on feasible solutions. The results of these experiments are provided in \cref{tab:demo_assignment_results}. 

\begin{table*}
% \tiny 
% \scriptsize
% \footnotesize 
% \small
\normalsize
\centering
\ra{1.11}
\caption{Comparison of task allocation methods. Values listed are predicted makespans in seconds. Entries with no data indicate no feasible solution was found in the allocated optimizer time limit.}
\begin{adjustbox}{max width=\textwidth}
\begin{threeparttable}
    \begin{tabular}{@{}llccc@{}}
        \toprule
        \multicolumn{2}{l}{Model (Parts$/$Assemblies)} & \multirow{2}{*}{\textsc{Greedy}\tnote{*}\phantom{}} & \multirow{2}{*}{\textsc{MILP}\tnote{$\dagger$}} & \multirow{2}{*}{$\textsc{MILP} + \textsc{Greedy}$\tnote{$\ddagger$}\phantom{a}} \\
        & \quad$\#$ Robots \\
        \midrule
        \multicolumn{2}{l}{Tractor $(20 / 8)$} \\
        & \quad$\phantom{0}\phantom{0}5$    & $27.4$ & $20.9$ & $20.8$  \\
        & \quad$\phantom{0}10$              & $16.1$ & $11.2$ & $11.1$  \\
        & \quad$\phantom{0}15$              & $11.9$ & $8.6$  & $8.6$  \\
        \multicolumn{2}{l}{X-Wing Mini $(61 / 12)$} \\
        & \quad$\phantom{0}15$              & $31.2$ & $23.4$ & $24.1$  \\
        & \quad$\phantom{0}20$              & $23.3$ & $18.5$ & $18.8$  \\
        & \quad$\phantom{0}25$              & $20.1$ & $17.4$ & $15.8$  \\
        \multicolumn{2}{l}{Imperial Shuttle $(84 / 5)$} \\
        & \quad$\phantom{0}15$              & $55.9$ & $-$ & $44.3$  \\
        & \quad$\phantom{0}20$              & $43.8$ & $33.4$ & $34.7$  \\
        & \quad$\phantom{0}25$              & $34.0$ & $27.4$ & $28.5$  \\
        \multicolumn{2}{l}{AT-TE Walker $(100 / 22)$} \\
        & \quad$\phantom{0}25$              & $37.1$ & $-$ & $30.2$  \\
        & \quad$\phantom{0}35$              & $29.9$ & $22.4$ & $22.7$  \\
        & \quad$\phantom{0}45$              & $23.4$ & $18.2$ & $18.5$  \\
        \multicolumn{2}{l}{X-Wing $(309 / 28)$} \\
        & \quad$\phantom{0}40$              & $144.5$ & $-$ & $134.5$  \\
        & \quad$\phantom{0}50$              & $124.8$ & $-$ & $113.7$  \\
        & \quad$\phantom{0}60$              & $106.2$ & $-$ & $99.2$  \\
        \multicolumn{2}{l}{Airplane $(326 / 28)$} \\
        & \quad$\phantom{0}40$              & $207.6$ & $-$ & $183.5$  \\
        & \quad$\phantom{0}50$              & $166.8$ & $-$ & $156.7$  \\
        & \quad$\phantom{0}60$              & $144.5$ & $-$ & $128.6$  \\
        \multicolumn{2}{l}{Star Destroyer $(418 / 11)$} \\
        & \quad$\phantom{0}65$              & $113.7$ & $-$ & $106.1$  \\
        & \quad$\phantom{0}75$              & $104.9$ & $-$ & $96.5$  \\
        & \quad$\phantom{0}85$              & $90.0$ & $-$ & $81.0$  \\
        \multicolumn{2}{l}{King's Castle $(761 / 70)$} \\
        & \quad$\phantom{0}75$              & $291.8$ & $-$ & $286.2$  \\
        & \quad$125$                        & $196.8$ & $-$ & $187.1$  \\
        & \quad$175$                        & $148.8$ & $-$ & $144.0$  \\
        \bottomrule
    \end{tabular}
    \begin{tablenotes}
        \footnotesize % Set table notes fontsize
        \item[*] {\greedyLEGO{}}
        \item[$\dagger$] {\SparseAdjacencyMILP{}}
        \item[$\ddagger$] {\SparseAdjacencyMILP{} with \greedyLEGO{} warm-start}
    \end{tablenotes}
\end{threeparttable}
\label{tab:demo_assignment_results}
\end{adjustbox}
\end{table*}

As expected, \SparseAdjacencyMILP{} outperformed \greedyLEGO{} when it was able to find a solution. However, the increased performance comes at the cost of an increase in computation time, especially for larger projects. Using the \greedyLEGO{} solution as a feasible warm-start for the \SparseAdjacencyMILP{} formulation improved the \greedyLEGO{} solution and was able to find an improved feasible solution in all experiments except for Saturn V. These results suggest that \greedyLEGO{} can provide quality solutions quickly, but when time allows, those solutions can be further refined when used as a feasible starting point for the \SparseAdjacencyMILP{} formulation using modern \textsc{MILP} solvers. 

\section{Discussion and Conclusion}\label{sec:LEGO_discussion}

As previously noted, our framework abstracts away many important details that would need to be considered in the real world. Here, we identify some of those considerations and point to the relevant literature and/or discuss how our approach could be extended in future work. 

Our current framework makes several simplifying assumptions that enable us to focus on the core coordination challenges of large-scale multi-robot assembly. We assume a homogeneous robot fleet with identical capabilities and load-carrying capacities, which simplifies task allocation and team formation. However, the underlying algorithmic structure is readily extensible to heterogeneous robot teams. The $\RobotGo{}$ nodes in our scheduling formulation (\cref{sec:collaborative_task_allocation}) can enforce role-specific constraints by restricting which robots may fulfill particular transport requirements. For example, heavy payloads could be constrained to robots with sufficient capacity, or specialized tasks could be assigned to robots with specific capabilities.

While this approach enables rapid planning by generating a single optimized construction plan, different designs often involve trade-offs between competing objectives such as spatial efficiency versus execution time, or robustness versus resource requirements. The modular structure of our algorithmic stack supports exploring such alternatives. For example, users can modify input constraints (buffer sizes, team configurations, collision tolerances) or optimization weights to generate multiple candidate designs and evaluate their performance across different metrics. Safety considerations could be emphasized by increasing staging area buffer sizes, implementing more conservative collision avoidance parameters, or designing redundant transport paths. Flexibility could be enhanced by maintaining spare robot capacity, designing modular staging layouts that can accommodate design changes, or implementing task allocation schemes that can quickly adapt to new requirements. Robustness to uncertainty and delays could be improved by building slack time into schedules, designing fault-tolerant task allocation that can handle robot failures, or implementing more conservative resource planning. Efficiency metrics such as robot utilization rates, energy consumption, or material handling distances could be optimized by modifying the objective functions in the task allocation formulation. While our current experiments focus solely on generation time, future work should incorporate these additional metrics to provide a more comprehensive performance assessment. 

Building on such multi-metric evaluations, another natural extension is to address the inverse design problem: given minimum acceptable thresholds for these metrics (e.g., makespan, throughput, or fault tolerance), determine the smallest number and types of robots required to satisfy those constraints. Although our current implementation does not explicitly solve this problem, the rapid solve times make it feasible to explore it in future work by running multiple optimization instances with varying team configurations to efficiently search for the minimal robot set meeting the specified criteria.

Beyond these coordination challenges, our framework does not address the fine manipulation and geometric path planning required to piece together complex assemblies. The closest we come to addressing this is to have each robot deposit each assembly component on the side of the staging area that is closest to the component's destination configuration within the assembly. We assume, rather, that these planning and control tasks are handled by some lower-level system. As noted in \cref{sec:LEGO_background}, existing work in multi-scale manipulation and collaborative grasp planning is particularly relevant in this regard \citep{Dogar2015,Dogar2019}.

Our method for configuring robot teams is based on a geometric heuristic. A more principled approach would consider factors like payload mass and mass distribution, structural properties, grasping locations, and the quality thereof. Existing work on multi-robot grasp planning offers a good starting point for the development of a more rigorous approach \citep{Muthusamy2015,Dogar2019,Tariq2018}.

Our staging plan layout procedure produces a global staging plan with at least one attractive property: each assembly can be transported in a straight path from its own staging area to its prescribed dropoff zone without crossing through other staging areas. However, deeply nested assemblies quickly lead to inefficient space usage. More space-efficient layouts could be achieved through approaches such as Voronoi diagrams that ``pull'' staging areas toward each other while maintaining necessary constraints. Additionally, our cylindrical staging zones can be wasteful for certain geometries (e.g., long, skinny assemblies would benefit from correspondingly shaped staging areas). The iterative layout approach could be generalized to other geometric shapes, with octagonal prisms being particularly promising since they can approximate both round and elongated parts. Another consideration is that our layout approach is based purely on the assembly specification and does not account for raw material storage locations or existing facility constraints, such as fixed component loading zones and machine placement. These practical constraints could be integrated into the spatial layout optimization with modest modifications to the constraint formulations.

Another important layout consideration is that our approach does not account for the temporal aspect of the assembly process. Staging areas only need to be separate from each other if they are being used simultaneously. In an environment with limited floor space, it might not be desirable or feasible to define a staging plan with no overlap between staging circles. Just as we allow deposit zones to overlap with the deposit zones from previous and future build steps, it would be useful to allow staging circles to overlap if their assembly construction timelines are far apart. Promising approaches to address these considerations might be found in the literature on facilities planning \citep{Tompkins2010}.

Our framework employs centralized planning followed by distributed execution, which enables global optimization but may face scalability limitations for very large robot fleets or dynamic environments with frequent disruptions. Extending to fully decentralized coordination, where robots make coalition formation and task allocation decisions locally, could improve scalability and fault tolerance, particularly for handling robot malfunctions or unexpected delays that were not explicitly considered in our current implementation.

Our three-layer distributed execution strategy works well in practice. In particular, the dispersion protocol is crucial to enable deadlock-free execution---our early experiments without the dispersion protocol (i.e, just \TangentBugPolicy+\RVO) were characterized by frequent deadlock due to crowding of inactive agents around the assembly staging areas. Task-swapping (\cref{sec:avoiding_deadlock}) prevents deadlock that might otherwise occur when an agent is blocked from reaching its carrying position by other robots that are participating in the same transport unit. This idea might be extended to a full online task allocation and coalition-forming approach (i.e., assign tasks and form teams on the fly, rather than making all assignments before beginning execution). The sit-and-wait subroutine avoids most undesirable ``dancing'' behavior, but could surely be replaced by a more elegant solution. 

It is important to note that we have not identified any theoretical guarantees on the performance of our execution strategy. Though our solution is effective for the projects considered, it may be vulnerable to edge cases that do not appear in our set of demo projects. Thus an important direction for future work is the development of continuous space-distributed execution strategies that are certifiably free from deadlocks. The concept of dynamic prioritization, which is present in the second and third layers of our distributed controller, is a promising starting point. Potential avenues for improvement include more sophisticated potential field methods, such as the method proposed by \cite{Fink2008} for decentralized multi-robot caging and pushing of planar objects. Another approach would involve the definition of virtual highways in which agents would be required to move, along with rules about when and where agents could enter and exit the highway. This line of work could build on existing research in automated guided vehicles \citep{Vis2006}. 

The application of multi-robot systems in the domain of assembly and manufacturing has the potential to revolutionize the speed, efficiency, and adaptability of the production process. We have presented a proof-of-concept system for multi-robot assembly planning. Given a project specification that specifies an assembly tree and a set of build phases, our algorithm is capable of synthesizing and executing construction plans involving assemblies with hundreds of parts. This process includes planning the carrying configurations of robot teams that will move objects and assemblies through the factory, setting up staging zones where each assembly will be incrementally pieced together, assigning robots to both solo and collaborative transport tasks, and enabling the robots to execute the staging plan in a distributed manner. Our main contribution is the sum total of these components. We feel that our work lays a solid foundation for future studies in the domain of multi-robot assembly systems.

\section*{Acknowledgments}
The authors would like to thank Ahmed Sadek, Mohammad Naghshvar, and the team at Qualcomm Corporate Research for their insightful feedback. This work was supported by Qualcomm, Siemens AG, and the National Science Foundation under grant No. DGE – 1656518.

\bibliographystyle{elsarticle-num-names}
\bibliography{cleaned_bib}

\end{document}